\documentclass{article} 
\usepackage{iclr2026_conference,times}

\usepackage{amsmath,amsfonts,bm}

\def\eqref#1{equation~\ref{#1}}

\def\1{\bm{1}}

\DeclareMathAlphabet{\mathsfit}{\encodingdefault}{\sfdefault}{m}{sl}
\SetMathAlphabet{\mathsfit}{bold}{\encodingdefault}{\sfdefault}{bx}{n}

\usepackage{hyperref}
\usepackage{url}

\usepackage{booktabs}       
\usepackage{nicefrac}       
\usepackage{microtype}      
\usepackage{xcolor}         
\usepackage{graphicx}
\usepackage{textcomp}
\usepackage{xcolor}
\usepackage{algorithm}
\usepackage{algpseudocode}
\usepackage{subfigure}
\newtheorem{theorem}{Theorem}[section]
\newtheorem{proof}{Proof}[section]

\title{Principal Components for Neural Network Initialization: A Novel Approach to Explainability and Efficiency}

\author{
  Thu Nguyen$^{1}$, Nhan Phan$^{2}$, Uyen Dang$^{2}$, Michael A. Riegler$^{1}$, Pål Halvorsen$^{1}$ \\
  $^{1}$Simula Metropolitan, Oslo, Norway \\
  $^{2}$University of Science, Vietnam National University Ho Chi Minh City, Vietnam \\
  \texttt{\{thu,michael,paalh\}@simula.no, \{nhanmath97,leuyendangrs\}@gmail.com}
}

\iclrfinalcopy 
\begin{document}

\maketitle

\begin{abstract}
Principal Component Analysis (PCA) is a commonly used tool for dimension reduction and denoising. Therefore, it is also widely used on the data prior to training a neural network. However, this approach can complicate the explanation of eXplainable Artificial Intelligence (XAI) methods for the decision of the model. In this work, we analyze the potential issues with this approach and propose Principal Components-based Initialization (\textbf{\textit{PCsInit}}), a strategy to incorporate PCA into the first layer of a neural network via initialization of the first layer in the network with the principal components, and its two variants PCsInit-Act and PCsInit-Sub. We will show that explanations using these strategies are more simple, direct and straightforward than using PCA prior to training a neural network on the principal components. We also show that the proposed techniques possess desirable theoretical properties.  Moreover, as will be illustrated in the experiments, such training strategies can also allow further improvement of training via backpropagation compared to training neural networks on principal components. 
\end{abstract}

\section{Introduction}\label{sec-intro}
Principal Component Analysis (PCA) is a widely used dimensionality reduction technique that transforms high-dimensional data into a lower-dimensional space while preserving as much variance as possible. By identifying the principal components, which are the orthogonal directions that capture the most variance in the data, PCA helps to eliminate redundancy, improve computational efficiency, and mitigate the effects of noise. It achieves this through eigenvalue decomposition of the covariance matrix or singular value decomposition (SVD) of the data matrix. 
In the context of neural network training, PCA offers significant advantages by decorrelating the input features, which in turn leads to a better-conditioned Hessian matrix. This is crucial because a poorly conditioned Hessian, often caused by correlated features, can result in slow convergence during gradient-based optimization, leading to inefficiencies in training \cite{montavon2012neural}. By transforming the input data into a set of orthogonal principal components, PCA ensures that gradient updates during backpropagation are more stable and effective, reducing the likelihood of oscillations or excessive adjustments along certain directions \cite{bishop2006pattern}. As a result, the optimization process becomes more efficient, with faster convergence toward the optimal solution. This makes PCA a valuable preprocessing technique for reducing the computational cost and enhancing the performance of neural networks, particularly in high-dimensional settings where the curse of dimensionality and correlated features are common \cite{hastie2009elements}. Consequently, incorporating PCA in the training pipeline can lead to significant improvements in both speed and accuracy for models trained on complex, high-dimensional datasets \cite{jolliffe2016principal}. Therefore, it is common to \textbf{\textit{train neural networks on the principal components (PCA-NN)}}.

However, this approach also has some limitations. When PCA is applied before a neural network (PCA-NN), perturbation-based XAI methods, such as LIME \cite{ribeiro2016should}, Occlusion \cite{zeiler2014visualizing}, and Feature Permutation \cite{breiman2001random}, are heavily impacted because they rely on modifying input features and observing their effect on predictions. This is because perturbing the input features instead of the principal components directly affects the SVD of the input or the covariance matrix of the input and principal components are linear combinations of multiple features.  Moreover, if perturbations are applied to principal components, the results are meaningful only in the transformed PCA space. Mapping the effects back to the original features requires using the PCA loading matrix, which defines the contribution of each original feature to each principal component. By redistributing the changes in principal components proportionally to their feature weights in the loading matrix, approximate feature-level insights can be obtained. However, this process introduces inaccuracies because the dimensionality reduction step in PCA discards some variance, and the transformation is not fully invertible. 
Overall, \textit{\textbf{perturbation-based XAI methods are inherently less interpretable in PCA-NN pipelines because the transformations obscure the direct relationship between input features and predictions}}. 

 Next, consider gradient-based XAI methods. For PCA-NN, the original input data \( X \) is transformed into a lower-dimensional representation \( Z = W^\top (X - \mu) \), where \( W \) is the PCA projection matrix and \( \mu \) is the mean of the input data, before being passed to the neural network. Gradient-based visualization methods such as saliency maps \cite{kadir2001saliency} and Grad-CAM \cite{selvaraju2017grad}, which rely on backpropagating gradients from the network’s output to its input, can still be applied in this setup by mapping the gradients from the PCA-transformed space \( Z \) back to the original input space \( X \). This mapping can be achieved using the chain rule, where \(\frac{\partial \text{output}}{\partial X} = \frac{\partial \text{output}}{\partial Z} \cdot \frac{\partial Z}{\partial X}\), and for PCA, the Jacobian \(\frac{\partial Z}{\partial X} = W^\top\). Consequently, gradients computed in the PCA space can be back-projected into the original input space to enable visualization. 
However, \textit{\textbf{PCA may remove or distort fine-grained input features that are crucial for effective gradient-based visualization, potentially reducing the interpretability and quality of methods like saliency maps or SmoothGrad}}. 

Next, while feature attribution methods like SHAP \cite{lundberg2017unified} and LIME \cite{ribeiro2016should} can still explain predictions in the PCA-transformed space, mapping these explanations back to the original feature space for explanation of the input requires leveraging the PCA loading matrix and comes with the limitation of approximation. 
Specifically, in a PCA-NN setup, SHAP can be applied directly to the PCA-transformed features \( Z \). To attribute importance to the original input features \( X \), the contributions to \( Z \) must be back-projected to \( X \) using the PCA matrix \( W \). However, \(W\) may not be an invertible matrix. Therefore, the explanation is not complete, but rather an approximation. A similar thing happens to LIME. In summary, these methods can attribute importance in PCA-NN systems, but \textbf{\textit{additional steps are required to map contributions from the reduced-dimensional PCA space back to the original input space, and the quality of these attributions depends on how much information is preserved by PCA}}.
 

These limitations motivate us to introduce \textbf{\textit{Principal Components based Initialization (PCsInit)}}, a novel initialization technique for the first layer of a neural network based on the principal components that is based on the similarity between the multiplication in PCA and the multiplication between neural network input and weight matrix. Putting PCA inside the neural network via initialization as in PCsInit allows the explanation using XAI methods to be the same as for a regular neural network without applying PCA.
Also, while PCA preprocessing requires maintaining and applying the transformation matrix during inference, PCsInit incorporates this information directly into the network weights, reducing operational complexity. Next, PCsInit allows the network to adaptively refine these initialized weights during training, potentially capturing more nuanced feature relationships that static PCA transformation might miss. Most importantly, PCsInit simplifies the explainability pipeline by eliminating the need to back-project through a separate PCA transformation step.
In addition, we also introduce PCsInit-Act, which applies an activation layer after the principal components to increase the neural network ability of nonlinear patterns. Moreover, we also introduce PCsInit-Sub, another variant of PCsInit that compute the principal components based on the subset of the input to increase the computational efficiency for large datasets. Note that these approaches are different from Neural PCA \cite{valpola2015neural}, which refers to using neural networks to approximate  PCA.

In summary, our contributions include (i) the introduction of PCsInit, a novel method that integrates PCA into neural networks via weight initialization; (ii) we propose two variants: PCsInit-Act, which enhances nonlinear pattern recognition by applying activation functions after the PCA-initialized layer, and PCsInit-Sub, which allows efficient scaling to large datasets by computing principal components on data subsets; (iii) a theoretical analysis demonstrating the desirable properties of the proposed approach regarding convergence and robustness under noisy data; and (iv) extensive experiments across multiple datasets confirming that PCsInit and its variants achieve performance comparable to or superior to PCA-NN while maintaining a more straightforward explainability framework.

\section{Methodologies}
In this section, we will introduce 
\textit{\textbf{Principal Components Analysis Initialization (PCsInit)}}, its two variations (PCsInit-Act and PCsInit-Sub), and relevant theoretical properties.  
The main idea of PCsInit can be described as follows: First, 
let $\mathbf{X} = [x_{ij}]$, where $i = 1,...,n; j = 1,...,p$, be an input data matrix of $n$ observations and $p$ features. Assume that the features are centered and scaled. Then, recall that the solution of PCA can be obtained using the singular value decomposition of $\mathbf{X}$:
    $\mathbf{X} = \mathbf{U}\mathbf{D}\mathbf{W}^T,$  
where $\mathbf{U}$ is an $n \times p$ orthogonal matrix, $\mathbf{W}$ is a $p \times p$ orthogonal matrix, and $\mathbf{D}$ is a $p \times p$ diagonal matrix with diagonal elements $d_1 \ge d_2 \ge ... \ge d_p \ge 0$. If $r$ eigenvalues are used, the projection matrix is $\mathbf{W}_r$, which consists of the first $r$ columns of $\mathbf{W}$. Then, the dimension-reduced version of $\mathbf{X}$ is $\mathbf{X}\mathbf{W}_r$.
This multiplication has some similarity with the multiplication between the input \(\boldsymbol{X}\) and the weight matrix \(\mathbf{W}\) of the first layer in a layer of a neural network. 
This motivates us to bring the PCA inside the neural network by simply initializing the first layer of a neural network with the principal components of the input and using no activation function for the first layer. The width of the first layer is chosen so that n\_components is retained or so that \(r\) percentage of the variance explained is retained. 


Formally, the training process of PCsInit is described in Algorithm \ref{alg-pcsinit}. In the first step, PCA is applied to the entire input dataset \( \boldsymbol{X} \) to extract the top \( r \) principal components, which are stored in the projection matrix \( \mathbf{W}_r \). These principal components represent the directions of maximum variance in the data, allowing for an efficient lower-dimensional representation while preserving critical information.  
In the second step, the matrix \( \mathbf{W}_r \) is used as the weight matrix for the first layer of a neural network \( f \). This initialization ensures that the first layer performs a transformation aligned with the most informative features of the input data, providing a strong inductive bias that can facilitate learning.  
At the third step, to stabilize training and allow the deeper layers to adapt to the PCA-based initialization, the first layer is frozen, meaning its weights remain unchanged, while the remaining layers of the network are trained for \( n_{\text{frozen}} \) epochs. This prevents drastic weight updates in the first layer, ensuring that the extracted principal components provide a meaningful starting point for training.  
At the fourth step, after the deeper layers have sufficiently adapted, the first layer is unfrozen, allowing the entire neural network to be trained jointly for an additional \( n \) epochs. This fine-tuning step enables all layers, including the first one, to update their parameters in a coordinated manner, optimizing the overall network performance and leading to better feature extraction and representation learning.
The purpose of the initial frozen phase and then fine-tuning is to allow the later layers to adapt to the PCA-based feature space, establishing stable higher-level representations. Once these layers have learned effective feature combinations, unfreezing the first layer enables fine-tuning of the initial PCA-based weights while maintaining the learned feature hierarchy. This two-phase approach prevents early disruption of the meaningful PCA structure while allowing eventual optimization of all parameters.

\begin{algorithm}[H]
\caption{\textbf{PCsInit process} }\label{alg-pcsinit}
\begin{algorithmic}[1]
    \State Applying PCA on the whole input \(\boldsymbol{X}\) to get the projection matrix \(\mathbf{W}_r\) that consists of \(r\) principal components,
    \State Use \(\mathbf{W}_r\) as the weight matrix for the first layer of the neural network \(f\),
    \State Freeze the first layer, and train the remaining ones for \(n_{frozen}\) epochs.
    \State Unfreeze the first layer and train the whole neural network for \(n\) more epochs.
\end{algorithmic}
\end{algorithm}
\vspace{-2mm}

\textbf{\textit{PCsInit versus PCA prior to neural network.}} The goal of \(\boldsymbol{X} \mathbf{w}_r\) is to find the principal directions (eigenvectors of the covariance matrix) that capture the most variance in the data, while for neural networks, the weight matrix \(\mathbf{W}\) is learned through backpropagation to minimize a task-specific loss function. In addition,  \(\mathbf{w}_r\) (the principal component directions) are computed directly from the data (using eigendecomposition or SVD) and are fixed once determined. Meanwhile, PCsInit for neural networks allows \(\mathbf{W}\) it to be adjusted iteratively during training for possible room for improvement.

Despite the differences, note that \textit{\textbf{for the PCsInit approach, if the weights of the first layer are frozen throughout the training process then its performance is the same as applying PCA to the input data and then training a model on the principal components}}. This is because the first layer can be seen as the PCA-conducting process and the subsequence layers act as the regular neural network layers. Even in this case, using PCA in the PCsInit manner with the first layer frozen during training makes it easier to explain the model. Specifically, a neural network built upon principal components makes decisions based on a transformed feature space where the most significant variations in the data are captured. Since PCA reduces dimensionality by keeping only the most informative features, the network learns patterns in terms of these principal components rather than the original raw features. The decision can be explained by analyzing which principal components contributed most to the output, mapping them back to the original features, and using techniques like SHAP or sensitivity analysis. Meanwhile, PCsInit allows using SHAP and other XAI techniques directly on the input. Therefore, \textit{\textbf{PCsInit offers a more straightforward explanation than explaining the decision of a neural network trained on principal components}}.

However, computing the principal components based on the whole input matrix can induce computational cost. Meanwhile, if the first layer is not completely frozen throughout training, it will be fine-tuned later. Therefore, it may be more computationally efficient to compute the principal components based on a subset of the input matrix instead. This motivates us to introduce \textbf{\textit{ PCsInit based on a SUBset of input data (PCsInit-Sub)}} as described in Algorithm \ref{alg-PCsInit-Sub}, which is a slight variation of PCsInit, where the principal components are computed based on a subset of the input matrix to reduce the computational expense associated with finding the principal components. 

\vspace{-2mm}
\begin{algorithm}[H]
\caption{\textbf{PCsInit-Sub process} }\label{alg-PCsInit-Sub}
\begin{algorithmic}[1]
    \State Applying PCA on a subset \(\boldsymbol{Z}\) of the input \(\boldsymbol{X}\) to get the projection matrix \(\mathbf{W}_r\) that consists of \(r\) principal components,
    \State Use \(\mathbf{W}_r\) as the weight matrix for the first layer of the neural network \(f\), 
    \State Freeze the first layer, and train the remaining ones for \(n_{frozen}\) epochs.
    \State Unfreeze the first layer and train the whole neural network for \(n\) more epochs.
\end{algorithmic}
\end{algorithm}
\vspace{-4mm}
While the first layer of PCsInit is similar to using PCA, the activation function is not applied after the first layer. Meanwhile, activation functions are known to be important in neural networks as they introduce non-linearity, allowing the model to learn complex patterns and relationships in data. In addition, they can also enable efficient gradient propagation during training. This motivates us to introduce \textbf{\textit{PCsInit with Activation function (PCsInit-Act)}}, another variation of PCsInit, which is described in Algorithm \ref{alg-PCsInit-Act}. It modifies PCsInit by applying an activation function after the first layer,
to improve the model's ability to capture non-linear patterns within the data. On the other hand, PCsInit-Sub focuses on computational efficiency by utilizing a subset of the input data instead of the entire dataset during initialization. As will be illustrated in the experiment section, this approach provides a lightweight yet effective alternative for handling large datasets, maintaining competitive accuracy and stability while reducing computational costs. Together, these variations improve the flexibility of the PCsInit methodology, offering improved interpretability and adaptability to diverse datasets and computational constraints.

\vspace{-2mm}
\begin{algorithm}[H]
\caption{\textbf{PCsInit-Act process} }\label{alg-PCsInit-Act}
\begin{algorithmic}[1]
    \State Applying PCA on the whole input \(\boldsymbol{X}\) to get the projection matrix \(\mathbf{W}_r\) that consists of \(r\) principal components,
    \State Use \(\mathbf{W}_r\) as the weight matrix for the first layer of the neural network \(f\),
    \State Add an activation function (ReLU, He,...) after the first layer used PCA-based initialization.
    \State Freeze the first layer, and train the remaining ones for \(n_{frozen}\) epochs.
    \State Unfreeze the first layer and train the whole neural network for \(n\) more epochs.
\end{algorithmic}
\end{algorithm}
\vspace{-2mm}

\vspace{-5mm}
\section{Theoretical analysis}\label{sec.theory}
\vspace{-3mm}
PCsInit acts as a form of structural regularization by injecting a strong, data-driven prior into the model. The prior is the assumption that the directions of highest variance in the input data (i.e., the principal components) are the most informative for the learning task. By initializing the first layer with these components, we are essentially guiding the network to first learn from the data's dominant, low-dimensional structure. Also, for noisy data, this initialization acts as a constraint, discouraging the model from immediately fitting noise. A standard, randomly initialized network might, by chance, align some of its initial projections with spurious, noisy features. PCsInit explicitly forces the network to start its learning process from a "denoised" and structurally meaningful representation of the data. Even after this layer is unfrozen, this strong initial bias can anchor the training process in a more robust region of the parameter space, improving generalization by reducing the risk of overfitting to noise.
In addition, while a rigorous analysis for deep, non-linear networks is intractable, we can build intuition by examining a simplified single-layer linear model. 
Without further comments, the norm used in this paper is Euclidean norm.

\subsection{Initialization analysis}  
PCA transforms the data into uncorrelated features. Therefore, the Hessian matrix (second derivative of the loss) becomes better-conditioned after PCA, making optimization more stable. While rigorous analysis for the whole neural network is challenging, we offer a simplifed case scenario to illustrate the performance of PCsInit in a simplified settings:
\begin{theorem}\label{theo-convergence-rate}
Consider a single-layer linear neural network regression model with input data $X \in \mathbb{R}^{d \times n}$ (where each column represents a data point), target labels ${Y} \in \mathbb{R}^{1 \times n}$, and weight vector $W \in \mathbb{R}^{d \times 1}$. Let $J(W) = \frac{1}{2} \|W^T X - {Y}\|^2$ be the Mean Squared Error (MSE) loss function, whose Hessian matrix with respect to $W$ is $H$. 
Suppose that $r$ principal components are used, $W_r$ is the matrix that consists of the selected eigenvectors of $X$, and let $Z = W_r^T X$. Then, 
the Hessian of $J(W_r)$  with respect to $W_r$ has a condition number $\kappa(H_r)$ that satisfies:
    $\kappa(H_r) \leq \kappa(H).$ 
\end{theorem}
Hence, in this case, PCsInit
leads to a better-conditioned optimization problem, potentially resulting in more stable and faster convergence of gradient-based optimization algorithms compared to training with the original data.
Now, recall that a function $f(x)$ is Lipschitz continuous if there exists a constant (Lipschitz constant) $L \geq 0$ such that for all $x, y$ in the domain of $f$: $ ||f(x) - f(y)|| \leq L ||x - y||$. $L$ represents the maximum rate at which the function can change. We have the following results regarding the Lipschitz constant for the first layer of PCsInit and PCsInit-Act:
\begin{theorem} \label{theo-Lipschitz}
Let $x \in \mathbb{R}^{d}$ be the input vector, and let $W_r \in \mathbb{R}^{d \times r}$ be the weight matrix of the first layer, where $r \leq d$ is the number of principal components used. The columns of $W_r$ are the principal component vectors. Assuming that the bias term is zero, i.e., the output of the first layer is $\qquad h^1 = W_r^T x$.  
Then, the Lipschitz constant for the first layer in PCsInit is:
 \(L_1 = \sigma_{max}(W_r) = ||W_r||.
    \)
\end{theorem}
\begin{theorem}\label{theo-lipschitz-PCsInit-Act}
Consider the first layer of a neural network with the weight matrix $W_r \in \mathbb{R}^{d \times r}$ consists of the first $r$ principal components (PCs) of the input data, forming orthonormal rows, such that $W_r = Q$, where $Q \in \mathbb{R}^{r \times d}$ has orthonormal rows ($Q Q^T = I_r$). Following this linear layer, an element-wise Lipschitz continuous activation function $\sigma: \mathbb{R} \rightarrow \mathbb{R}$ with Lipschitz constant $L_\sigma$ is applied. Then, the Lipschitz constant $L_1$ of the first layer operation $f_1(x) = \sigma(W_r^T x)$ with respect to the $L_2$ norm is $L_1 = L_\sigma$.
\end{theorem}
Theorem \ref{theo-Lipschitz} shows that the Lipschitz constant of the first layer in PCsInit is equal to the largest singular value  of the weight matrix $W_r$, which is formed by the principal component vectors. This means that the maximum amount by which the output of the first layer can change for a unit change in the input is given by the largest singular value of $W_r$. Next, Theorem \ref{theo-lipschitz-PCsInit-Act} shows that the Lipschitz constant of the first layer operation is simply the Lipschitz constant of the activation function used.

Now consider the Lipschitz constant of the first layer of a neural network based on the initialization method employed. PCsInit exhibits a fixed Lipschitz constant of $||W_r||$. Since $||W_r||$ is commonly orthonormal in implementation, this means  the Lipschitz constant of PCsInit is a constant $||W_r||=1$, a characteristic it shares with orthogonal initialization \cite{mishkin2015all}, which initializes weight matrices to be orthogonal and thus also possesses a constant Lipschitz constant. This inherent consistency can contribute to enhanced stability within the first layer. In contrast, He and Xavier initializations result in random Lipschitz constants that differ across initializations, introducing variability that can influence training dynamics. 

\subsection{Robustness to noisy data}

Note that PCA decorrelates the input features. If the noise is independent across the original input features, then PCsInit also decorrelates the noise in the output of the first layer. Meanwhile, He and Xavier initialization can create correlations in the noise, as they involve random linear combinations of the input dimensions. Moreover, note that while scaling down the norm of any standard initializer can help denoising, it reduces the magnitude of both the signal and the noise indiscriminately. This can be detrimental, as suppressing the signal too much can lead to vanishing gradients and hinder the learning process. In contrast, PCsInit provides a principled, data-driven structure. The first layer is initialized with principal components, which are the directions of maximum variance in the data. This means that the initialization is explicitly designed to preserve the signal's most important structural information while discarding dimensions with low variance, which are more likely to be dominated by noise. Therefore, PCsInit aims to maximize the signal-to-noise ratio (SNR) in the first layer's output, whereas simply reducing the norm of a random matrix may decrease the overall magnitude but can also worsen the SNR. In addition, we present the following statements regarding the robustness of PCsInit to noisy data, which illustrates that PCA projects the input noise onto the principal subspace. The variance of the noise in each principal component direction is scaled by the corresponding eigenvalue of $X^T X$. 
\vspace{-3mm}
\begin{theorem}\label{theo-robustness}
Assume that $\tilde{x} = x + \eta,$ where $\tilde{x}$ is the noisy input, $x$ is the clean input, and $\eta$ is the noise vector. In addition, assume also that $\eta \sim \mathcal{N}(0, \sigma^2I)$, i.e., the noise follows a Gaussian distribution with zero mean and covariance matrix $\sigma^2I$. Here, $\sigma^2 \in \mathbb{R}^+$ and $I$ is the identity matrix. 
Next, let the eigenvalues of $X^T X$ be $\lambda_1, \lambda_2, ..., \lambda_r$.
Then, for PCsInit, the noise propagated after the first layer is $W_r^T \eta$ 
follows Gaussian distribution with mean 0 and its covariance matrix is a diagonal matrix with entries $\sigma^2 \lambda_1, \sigma^2 \lambda_2, ..., \sigma^2 \lambda_r$.
\end{theorem}
Consider other initialization schemes for the first layer, if the input is noisy ($\tilde{x} = x + \eta$), the output of the first layer is $W_1\tilde{x} = W_1x + W_1\eta$. The noise component is $W_1\eta$. So, He and Xavier initializations transform noise by multiplying it with a randomly initialized matrix with a variance scaling. Meanwhile, orthogonal initialization  rotates the noise vector without changing its norm.
Next, we have the following theorem regarding the bound for the noise.
\begin{theorem}
    Consider a neural network where the first layer's weight matrix, \(W_r \in \mathbb{R}^{d\times r}\), is initialized as in PCsInit, and therefore \(W_r = Q\), where \(Q \in \mathbb{R}^{d \times r}\) is orthonormal. 
     Suppose that the input to the network is corrupted by additive white noise, i.e.,  \(\tilde{x} = x + \eta\), where \(x \in \mathbb{R}^{d}\) is the clean input signal and \(\eta \sim \mathcal{N}(0, \sigma^2 I)\) is the additive white noise.
     Then, the norm of noise component after the first layer is preserved, i.e.,
        \(   ||W_r^T\eta|| = ||\eta||.   \)
     Also, let \(f: \mathbb{R}^{d} \rightarrow \mathbb{R}^{r}\) be the neural network function, decomposed into layers \(f = f_L \circ f_{L-1} \circ ... \circ f_1\), where \(f_i\) represents the \(i\)-th layer,
     \(f(x)\) is the clean output, and \(\tilde{f}(\tilde{x})\) is the noisy output.
    Further assume that each subsequent layer \(f_i\) for \(i = 2, ..., L\) is \(L_i\)-Lipschitz continuous. Then,
        \(   ||\tilde{f}(\tilde{x}) - f(x)|| \le \left( \prod_{i=2}^{L} L_i \right) ||\eta||.   \)
\end{theorem}
While the norm of the noise in the first layer is preserved for PCsInit, for other initialization methods (e.g., He, Xavier) where the first layer's weight matrix \(W'\) does not have orthonormal rows, \(||W' \eta||\) is not guaranteed to be equal to \(||\eta||\) and can be larger or smaller. In fact, we have \( ||W' \eta|| \le ||W'|| ||\eta||,\) and it is possible that \(||W'|| > 1\). So, the noise can be amplified in the first layer.
Next, we have the following result, which provide deeper details regarding the noise of layer 2 and subsequent layers:
\begin{theorem}
For any layer $\ell > 1$, let $\rho^{\ell}$ is the activation function, $W^{\ell}$ is the weight matrix, and $b^{\ell}$ is the bias vector. Then, the output at layer $\ell$ is: $h^{\ell} = \rho^{\ell}(W^{\ell} h^{\ell-1} + b^{\ell}).$ Next, let $\eta^{\ell}$ represents the noise in the output of layer $\ell$ and assume also that $\rho^{\ell}$ is Lipschitz continuous with Lipschitz constant $L_{\ell}$, i.e.
$||h^{\ell} - \tilde{h}^{\ell}|| \leq L_{\ell} ||W^{\ell}|| \cdot ||\eta^{\ell-1}||,$
where $\tilde{h}^{\ell}$ is the clean output and $h^{\ell}$ is the noisy output.
Then, the bound for the noise propagated to the second layer is
    \(\qquad ||\eta^2|| \leq L_2 ||W^2|| \cdot ||W^1|| \cdot ||\eta^0||.\)
In addition, the general noise bound for any layer $\ell > 1$ is:
\(||\eta^{\ell}|| \leq \left[\prod_{i=2}^l(L_i||W^i||)\right]\cdot ||W^1|| \cdot ||\eta^0||.\)
\end{theorem}
This theorem provides a upper bound on how input noise propagates through the network, and shows that the noise at any layer l is directly proportional to $||W^1||$, the norm of the first layer's weight matrix. For PCsInit, ($||W^1|| = 1)$. The first layer, therefore, does not amplify the input noise. Meanwhile With He or Xavier initialization, $W^1$ is a random matrix whose norm is not strictly controlled. Depending on the dimensions and the specific random draw, its norm can be greater than 1, leading to an initial amplification of noise. By guaranteeing $||W^1|| = 1$, PCsInit prevents a potential explosion of noise magnitude at the very first layer. This is crucial because any amplification in the first layer will be propagated and potentially magnified through all subsequent layers. 
\vspace{-3mm}
\section{Experiments}\label{sec-experiment}

\subsection{Experiment Settings}\label{subsec-experiment-settings}

We conduct experiments on seven datasets with various characteristics whose details are found in Table \ref{table_info_datasets}. Among them, Parkinson and Micromass~\cite{Dua:2019} have the number of features significantly higher than the number of samples, MNIST \cite{deng2012mnist} and CIFAR10 \cite{krizhevsky2009learning} are image data, and HTAD \cite{garcia2021htad} is a noisy sensor-collected dataset. 
To demonstrate the effectiveness of our proposed initialization strategies, we compare them with
PCA-NN. For PCsInit, the first layer weights were initialized using principal components derived from the training dataset, while the first layer of the standard NN was initialized using the specified initialization technique (He, Xavier, or Orthogonal initialization). Subsequent layers in both PCsInit and the standard NN, as well as all layers of the PCA-NN (which operated on pre-computed principal components), also employed one of three initialization techniques. For PCsInit-Sub, the initial PCA was performed on a randomly selected 20\% subset of the training data. For PCsInit-Act, the ReLU activation function is used for the first layer.
In the PCsInit family, the first layer is initialized with the principal components, and then it is frozen for the first 30 epochs. After that, we unfreeze this first layer and train the whole model for 170 more epochs. To facilitate a fair comparison, for all strategies, all layers except the first layers of the PCsInit family are initialized with the same weights. The first layer of PCsInit and its variants are initialized based on principal components. 
To compare the explainability of the proposed methods against PCA-NN, 
we employ the Shapley Additive Explanations (SHAP) framework. Kernel SHAP calculates feature importance based on Shapley values from game theory combined with coefficients from a special weighted linear regression, allowing it to be applied to any model \cite{lundberg2017unified}. More details are in the Appendix.

\vspace{-3mm}
\subsection{Results and analysis}
Due to the page limit, the results of performance loss and accuracy is reported in figures in Appendix A.2. The results illustrate the performances of the proposed techniques against PCA-NN.
Notably, the PCsInit-Act variation shows strong performance on the given datasets.
In addition, when dealing with noisy datasets, PCsInit family generally performed better than PCA-NN. For instance, on the HTAD dataset, a noisy dataset, PCsInit and its variations were particularly effective. They were often the most accurate, learned quickly and steadily, and performed better than both PCA-NN. Similarly, on the MNIST dataset with added Gaussian noise, the PCsInit-Act variation performed much like the other PCsInit variations. 
For datasets with a large number of features compared to the sample size, such as Micromass (figure \ref{fig-micromass-he}) and Parkinson (figure \ref{fig-parkinson-he}), the PCsInit family gives pronounced improvement compared to  PCA-NN. On Micromass, while standard PCsInit still well, its variations (PCsInit-Act and PCsInit-Sub) offer enhanced stability and competitive accuracy.

\begin{figure}[ht]
    \centering
    \begin{minipage}[b]{0.47\linewidth}
        \centering
        \includegraphics[width=\linewidth]{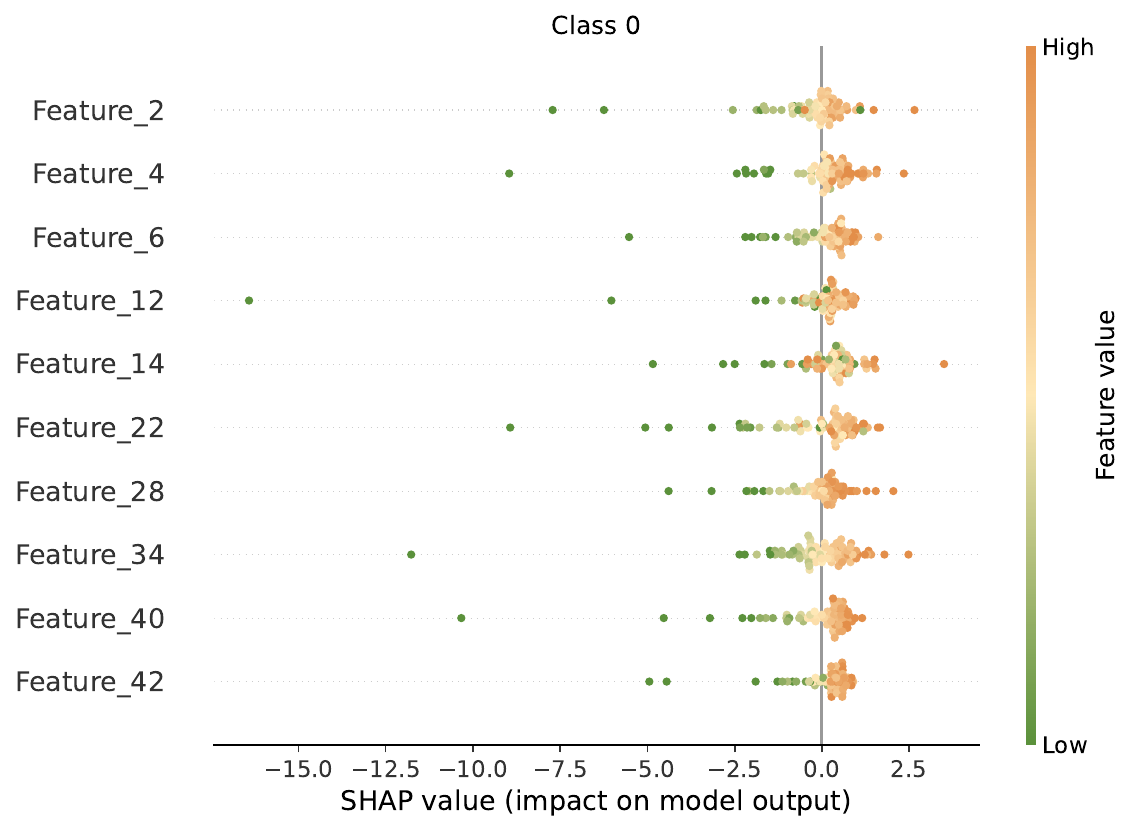}
        \label{fig-PCsInit-FC-class0}
    \end{minipage}
    \hfill
    \begin{minipage}[b]{0.47\linewidth}
        \centering
        \includegraphics[width=\linewidth]{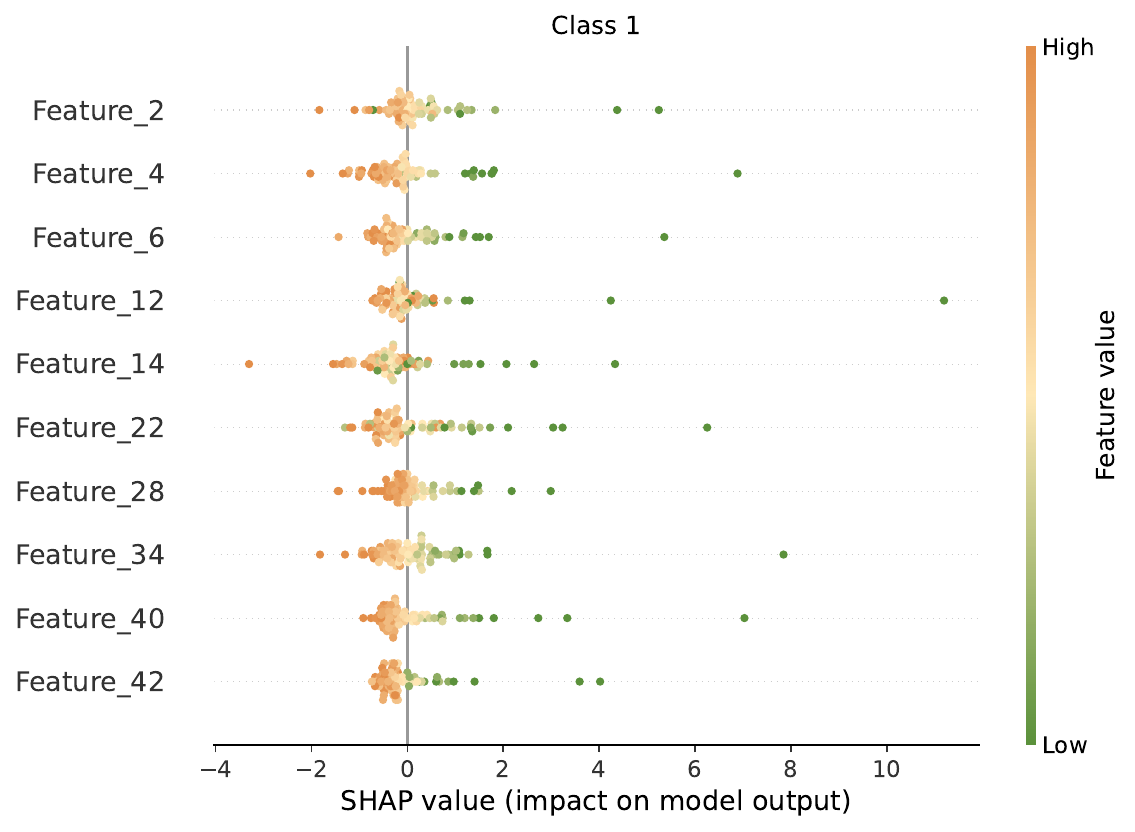}
        \label{fig-PCsInit-FC-class1}
    \end{minipage}
    \vspace{-4mm}
    \caption{Top 10 most important features contributing to \textbf{PCsInit} predictions on the Heart dataset for Class~0 and Class~1 across the test set.}
    \label{fig-PCsInit-FC}
\end{figure}

\begin{figure}[!ht]
    \centering
    \begin{minipage}[b]{0.47\linewidth}
        \centering
        \includegraphics[width=\linewidth]{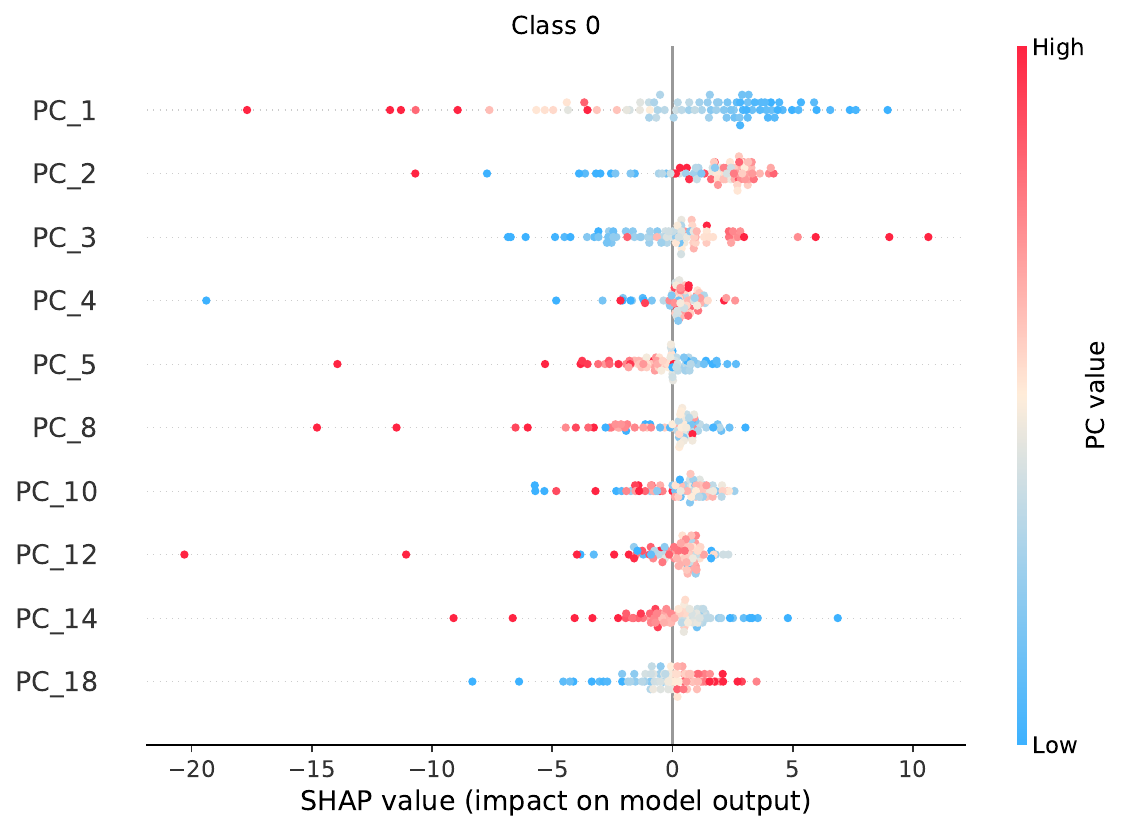}
        \label{fig-FFW-PC-class0}
    \end{minipage}
    \hfill
    \begin{minipage}[b]{0.47\linewidth}
        \centering
        \includegraphics[width=\linewidth]{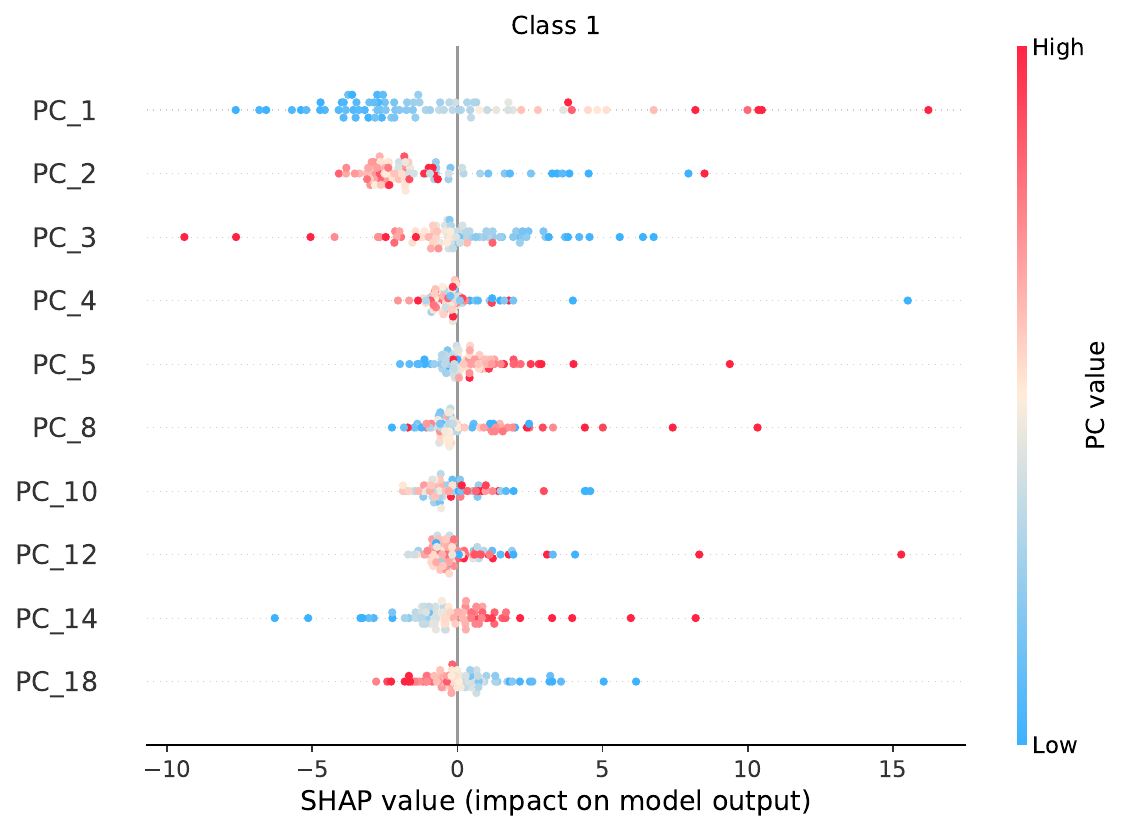}
        \label{fig-FFW-PC-class1}
    \end{minipage}
    \vspace{-4mm}
    \caption{Top 10 most important principal components contributing to \textbf{PCA-NN} predictions on the Heart dataset for Class~0 and Class~1 across the test set.}
    \label{fig-FFW-PC}
\end{figure}

Regarding interpretability, an illustration can be seen in figures \ref{fig-PCsInit-FC} and \ref{fig-FFW-PC}.
For PCsInit, feature contributions can be computed directly by using the Kernel SHAP method, whereas in the PCA-NN method, they can only be approximated by first calculating SHAP values of the principal components and then deriving feature importance based on the contribution of each feature to those components.

\begin{figure}[!ht]
    \centering
    \includegraphics[width=\linewidth]{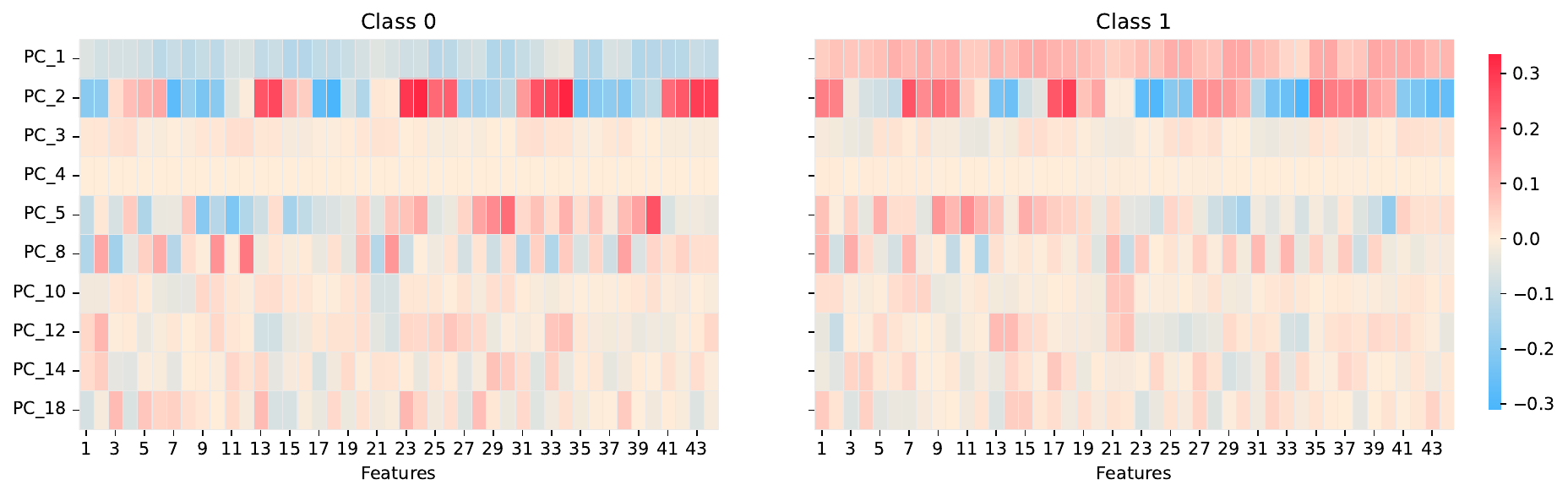}
    \vspace{-4mm}
    \caption{Heatmap of feature importance through 10 important principal components for the \textbf{PCA-NN} predictions on the Heart dataset for Class~0 and Class~1.}
    \label{fig-FFW-feature}
\end{figure}

\textbf{Global Feature Importance}: Figure~\ref{fig-PCsInit-FC}, which presents the global feature importance obtained using the PCsInit method, shows that it is straightforward to identify the features that directly contribute to the predictions, with the top 10 features exhibiting the strongest influence. In contrast, Figure~\ref{fig-FFW-PC} shows the global principal component importance in the PCA-NN method, which highlights the contribution of principal components rather than original features. To further interpret the feature contribution, the heatmap in Figure~\ref{fig-FFW-feature} illustrates the importance of each original feature through each principal component. For instance, in Class~0, features indexed by 7, 17, 23, 24, and 34 strongly influence predictions through principal component 2 (PC\_2), as indicated by the more intense colors. However, determining how each feature impacts  the model's predictions in this case requires more effort and further analysis.

\begin{figure}[!ht]
    \centering
    \begin{minipage}[b]{0.47\linewidth}
        \centering
        \includegraphics[width=\linewidth]{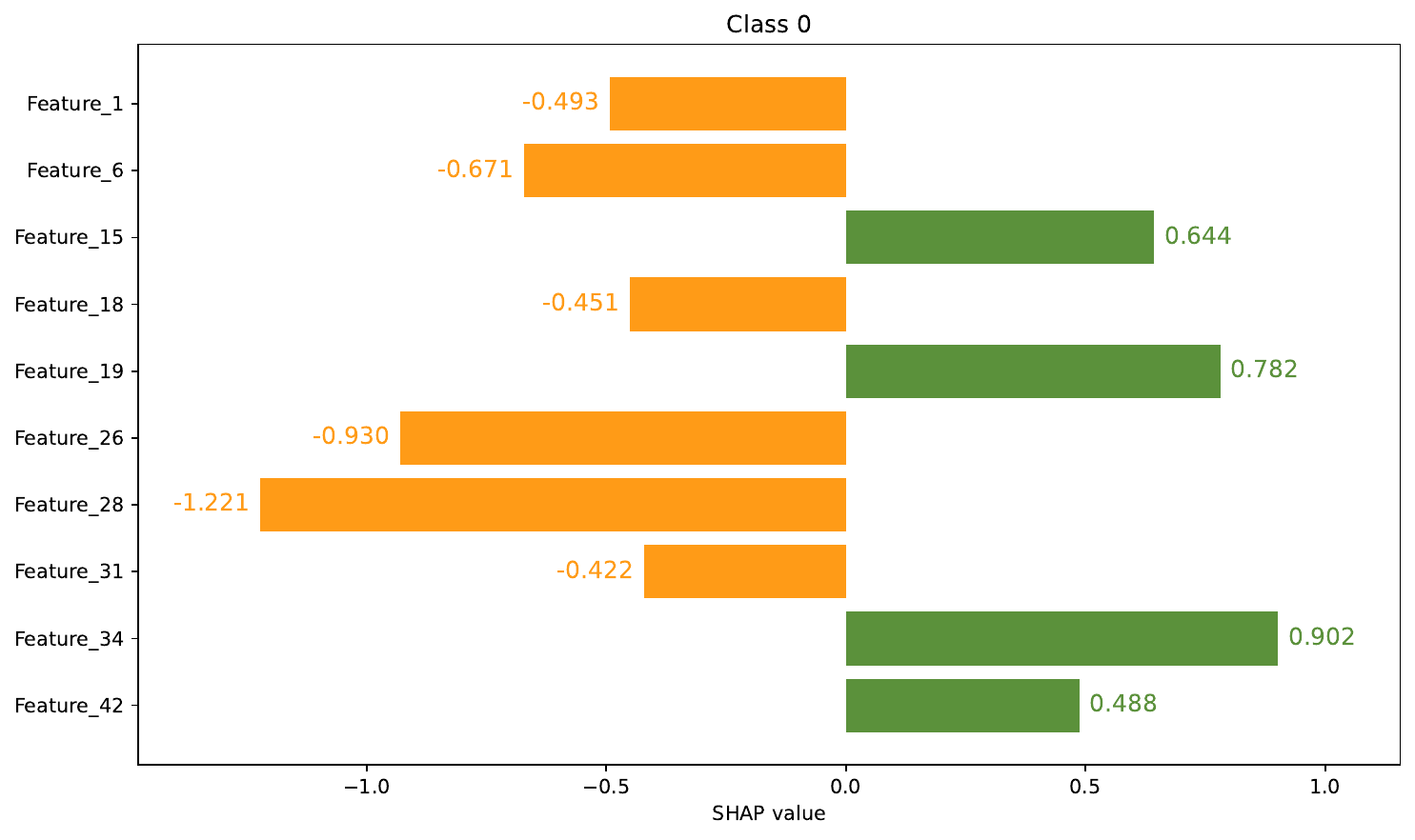}
        \label{fig-PCsInit-local-class0}
    \end{minipage}
    \hfill
    \begin{minipage}[b]{0.47\linewidth}
        \centering
        \includegraphics[width=\linewidth]{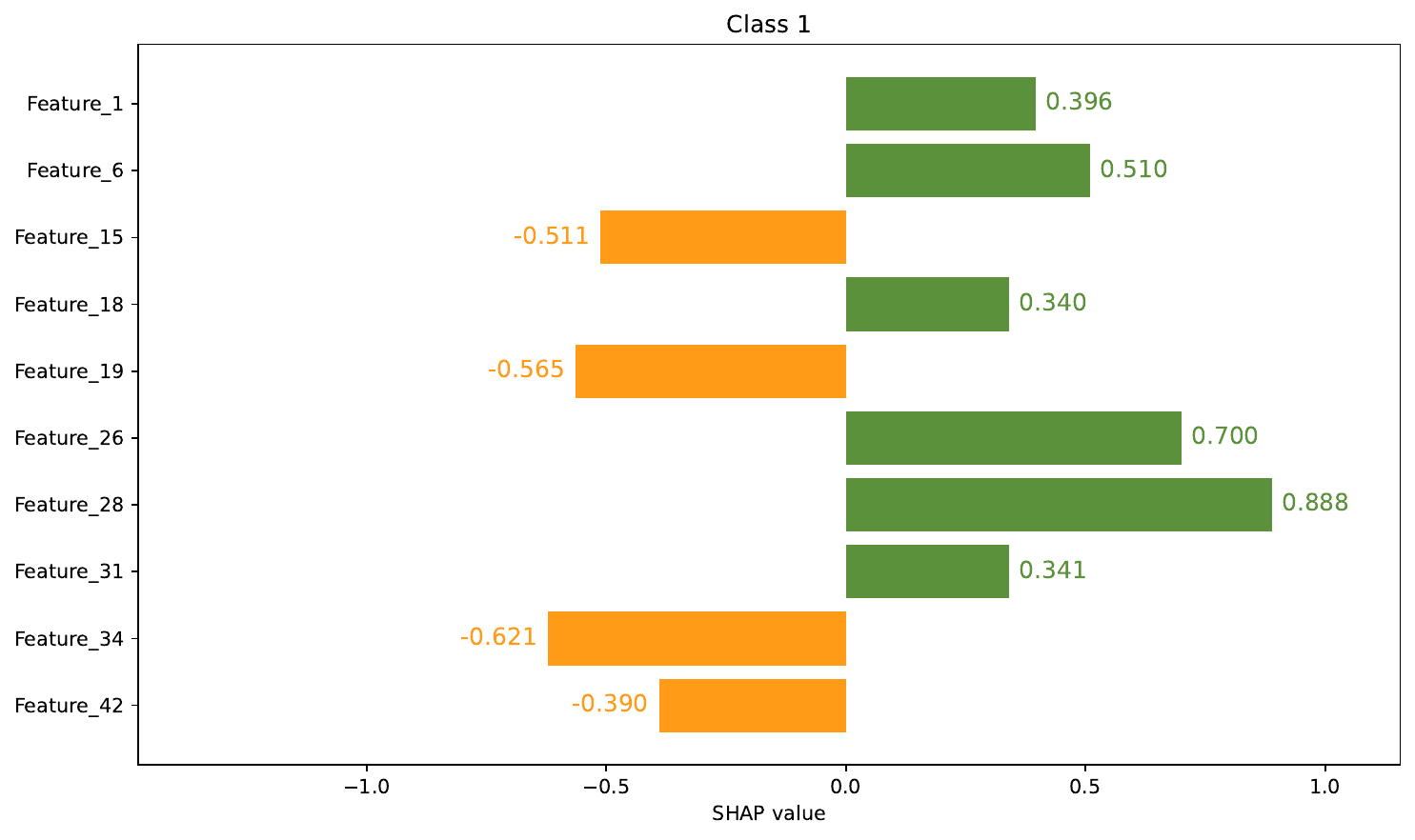}
        \label{fig-PCsInit-local-class1}
    \end{minipage}
    \vspace{-4mm}
    \caption{The local feature importance of a data point contributing to the \textbf{PCsInit} prediction on the Heart dataset for Class~0 and Class~1.}
    \label{fig-PCsInit-local}
\end{figure}

\begin{figure}[!ht]
    \centering
    \begin{minipage}[b]{0.47\linewidth}
        \centering
        \includegraphics[width=\linewidth]{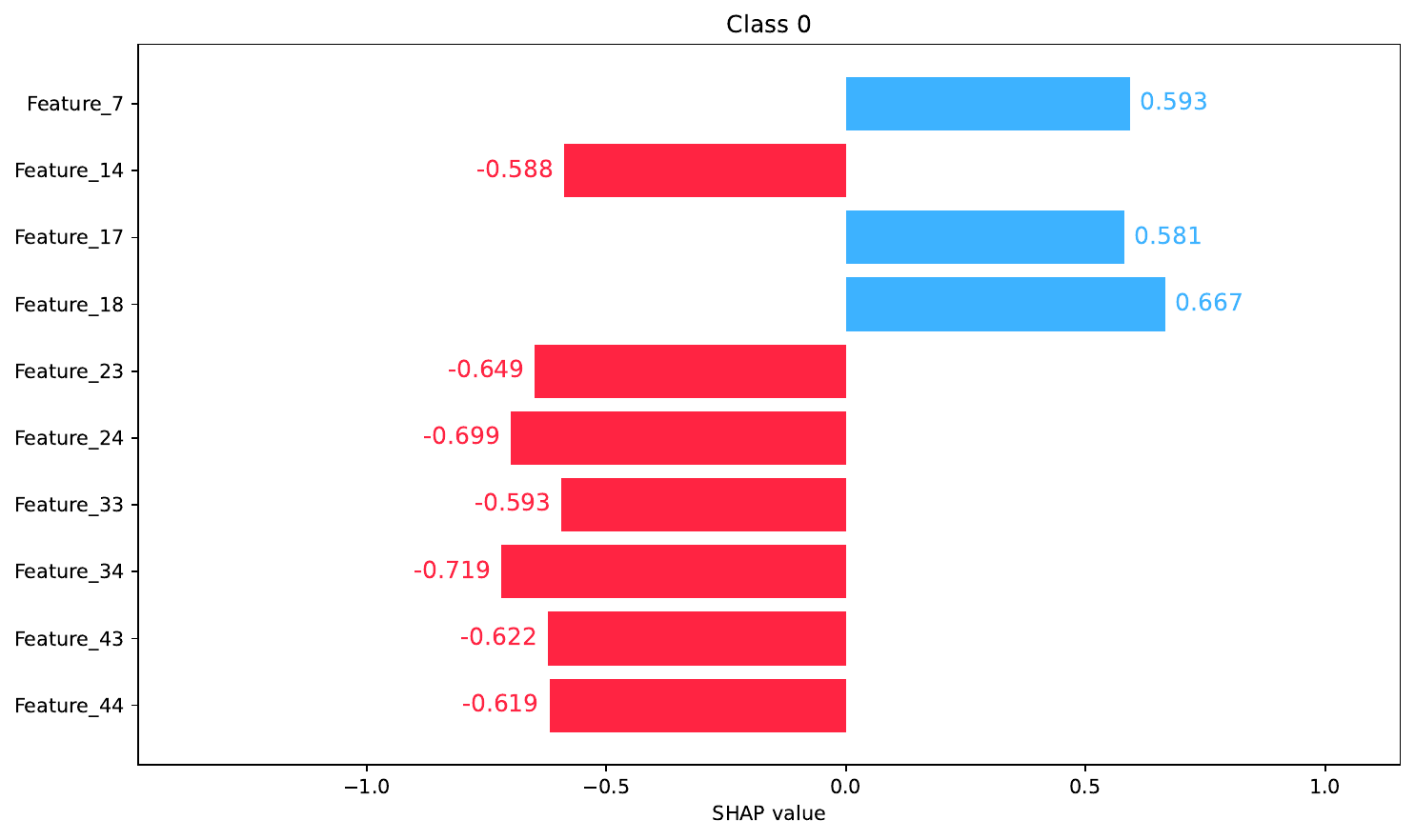}
        \label{fig-PCANN-local-class0}
    \end{minipage}
    \hfill
    \begin{minipage}[b]{0.47\linewidth}
        \centering
        \includegraphics[width=\linewidth]{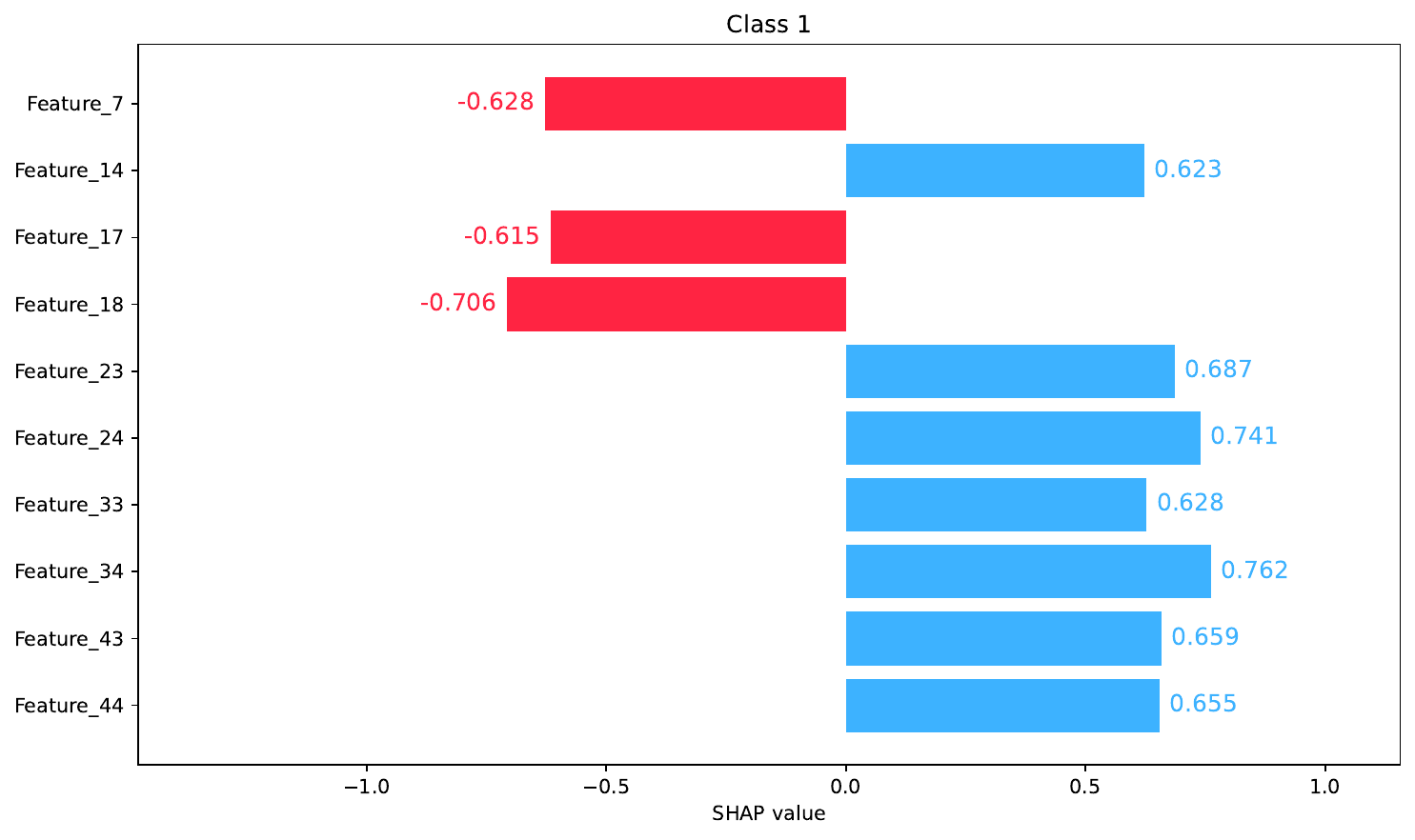}
        \label{fig-PCANN-local-class1}
    \end{minipage}
    \vspace{-4mm}
    \caption{The local feature importance of a data point contributing to the  \textbf{PCA-NN} prediction through PC\_2 on the Heart dataset for Class~0 and Class~1.}
    \label{fig-PCANN-local}
\end{figure}

\textbf{Local Feature Importance}:
Figure \ref{fig-PCsInit-local} shows the SHAP values for a sample point using PCsInit. Here, feature\_28 has the strongest influence on the outcome for both Class~0 and Class~1. Since PCsInit directly computes feature contributions without projection, it allows for identifying the driving factors behind individual predictions easily.
By contrast, Figure \ref{fig-PCANN-local} illustrates the approximate SHAP values for the same sample point through PC\_2 using the PCA-NN method. It should be noted that across different principal components, the same features may show different SHAP values. This makes two layers of interpretation: first, feature contributions are dispersed across multiple components; second, each component contributes to the final prediction with a different weight. As a result, the relationship between original features and the final predictions is indirect and more difficult to trace.

In general, the above analysis demonstrates that our PCsInit method provides more direct, transparent, and stable explanations, while PCA-based methods increase the complexity in model interpretability.

Regarding running time, consider Figure \ref{fig-time-heart-he}, one can see that PCsInit-Sub proved to improve the running time by computing the principal components on a subset of the data. However, as the number of the epochs increase, the PCsInit family has higher cumulative running time compared to PCA-NN. This is because for PCA-NN, the input is the principle components, and therefore have less features. Meanwhile, for the PCsInit family, the input size is the same as the original input, as the network is intialized with PCA instead of being trained on the pricipal components.

\vspace{-5mm}
\section{Conclusion}\label{sec-concl}
In this work, we introduced PCsInit, a novel neural network initialization technique designed to embed Principal Component Analysis (PCA) within the network's first layer.  A key advantage of PCsInit, along with its variants PCsInit-Sub and PCsInit-Act, is the enhanced clarity and directness it brings to model explainability, avoiding the complexities associated with applying PCA as a preprocessing step.   The theoretical properties of PCsInit, particularly its positive impact on Hessian conditioning, further solidify its value.

However, a draw back of PCsInit is that requires computing PCA on the entire dataset, which can computationally expensive for high-dimensional data (Though, it is also worth to mention that PCA-NN also requires computing PCA on the entire dataset). PCsInit-Sub is a more scalable approach as it requires computing PCA only on a subset of the data, without sacrificing the accuracy, as illustrated in the experiments. Another drawbacks of PCsInit roots from the fact that it use PCA, which may lead to loss in spatial information if compared to standard neural network without PCA (however, PCA-NN also suffer the same issue). This means that PCsInit may not be suitable for convolutional neural networks. Therefore, to mitigate this issue, we will consider dimension reduction techniques that keeps spatial information such as SpatialPCA, for example.
Moreover, our future research will delve into the broader applicability of PCsInit across different network architectures and data characteristics, including the exploration of Kernel PCA-based initialization. In addition, we will examine the performance of the proposed approaches under various neural network architectures and various types of data (noisy, imbalanced). In addition, we will explore the potential usage of Kernel PCA instead of PCA for initializing neural networks, as well as analyze in depth the choice of initialization techniques for the subsequent layers after the first layer for the PCsInit family.
\bibliographystyle{iclr2026_conference}
\bibliography{ref}

\appendix
\section{Appendix}
\subsection{Dataset and experiments description}
For all datasets except MMIST, the experiments are repeated 10 times with 70\%/30\% train/test split ratio, and the average is reported in the graphs. For the MNIST dataset, we simply used the already split train-test portions. Besides, Gaussian noise (mean 0, standard deviation 1) was added to each pixel feature after the initial standardization of the input data. 

For PCA, we choose the number of principal components so that a minimum of 95\% of variance is retained, as commonly used in various research articles \cite{nguyen2022principle,audigier2016multiple,do2024blockwise,nguyen2023principal}. The width of the network used for each dataset has a width equal to the number of principal components retained. All models are trained with a total of 200 epochs using Cross Entropy Loss and Adam optimizer.  
The number of layers used for all datasets is 5.

The experiments were run on a CPU of AMD Ryzen 3 3100 4-Core Processor, 16.0 GB RAM. The codes for the experiments will be made available upon acceptance. 
\begin{table}[htbp]
\caption{Descriptions of datasets used in the experiments}
\begin{center}
    \begin{tabular}{|c|c|c|c|}
		\hline
		Dataset & \# Classes & \# Features & \# Samples \\
		\hline
		Heart & $2$ & $44$ & $267$ \\\hline
		Ionosphere & $2$ &  $32$ &  $351$\\\hline
        Parkinson & $2$ & $754$ & $756$ \\\hline
		Micromass  & $10$ & $1087$ & $360$ \\\hline
        HTAD & $7$& $54$& $1386$\\\hline
        MNIST & $10$ & $784$ & $60000$ \\\hline
	\end{tabular}
	\label{table_info_datasets}
\end{center}
\end{table}

\subsection{Experiment results}\label{apd-res-He}

The experimental results depicted in the figures demonstrate that PCsInit variations outperform the PCA-NN initialization across most datasets and initialization techniques.

\begin{figure}[!ht]
    \centering
    \includegraphics[width=0.7\linewidth]{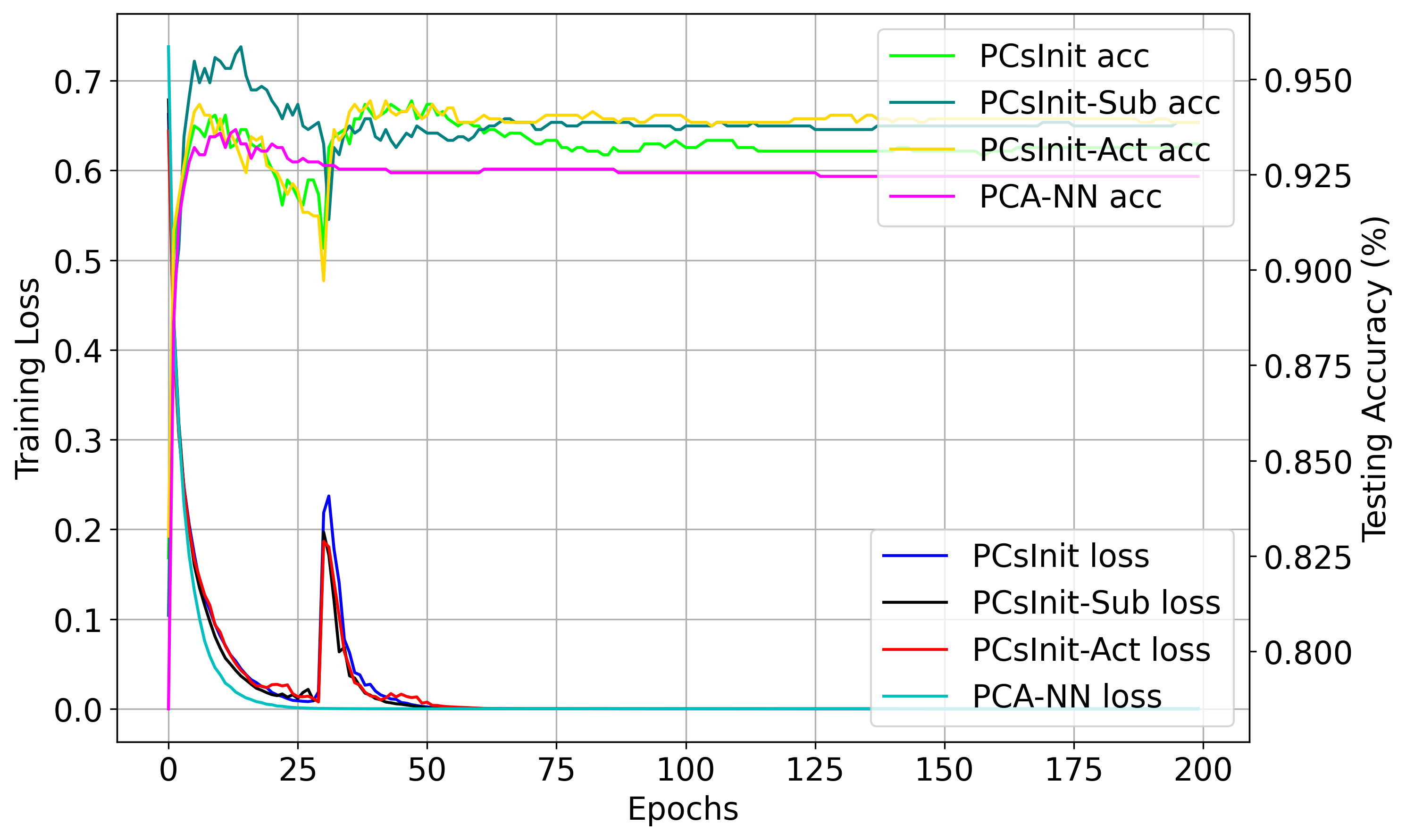}
    \vspace{-4mm}
    \caption{Ionosphere dataset with He initialization}
    \label{fig-ionosphere-he}
\end{figure}

\begin{figure}[!ht]
    \centering
    \includegraphics[width=0.7\linewidth]{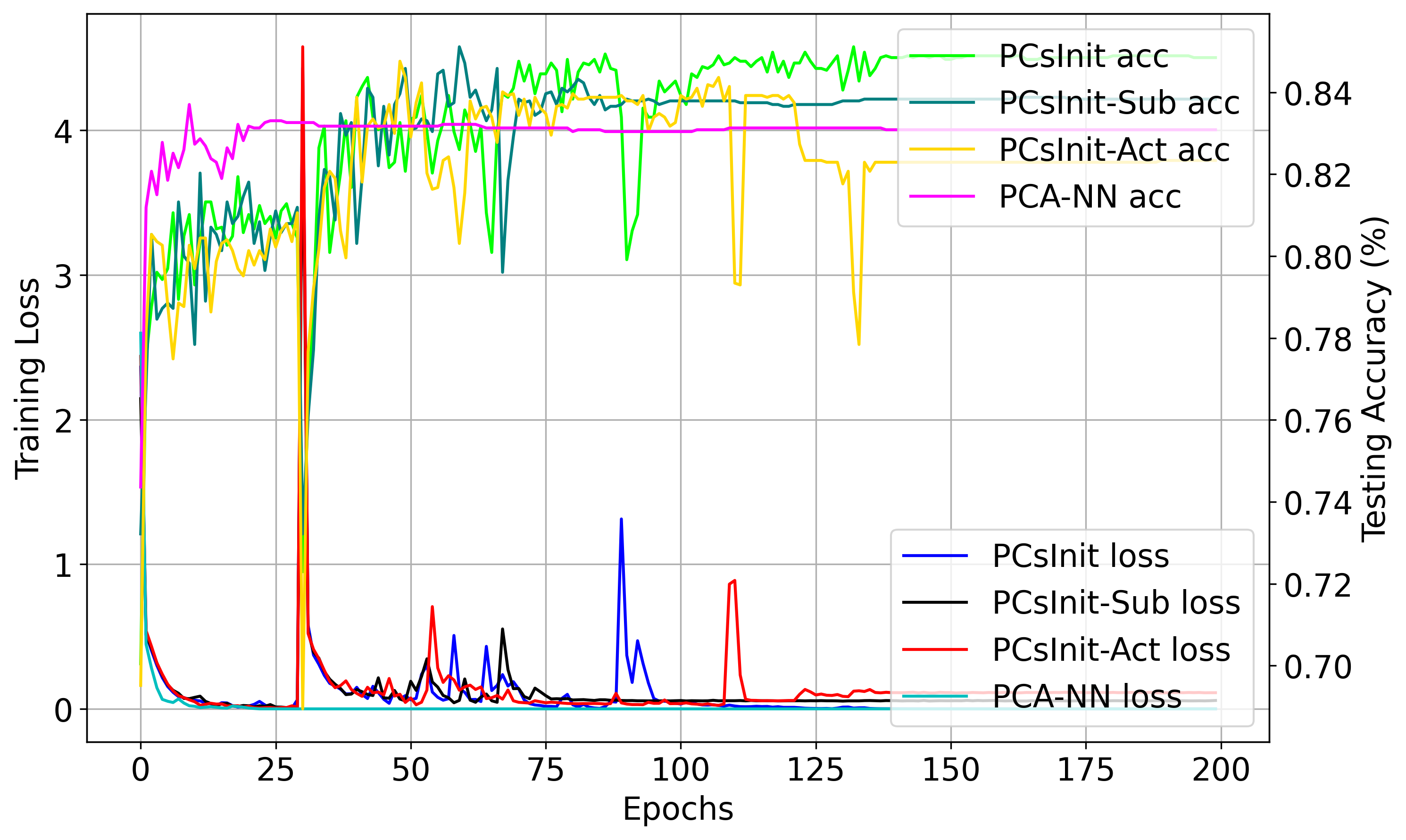}
    \vspace{-4mm}
    \caption{Parkinson dataset with He initialization}
    \label{fig-parkinson-he}
\end{figure}

\begin{figure}[!ht]
    \centering
    \includegraphics[width=0.7\linewidth]{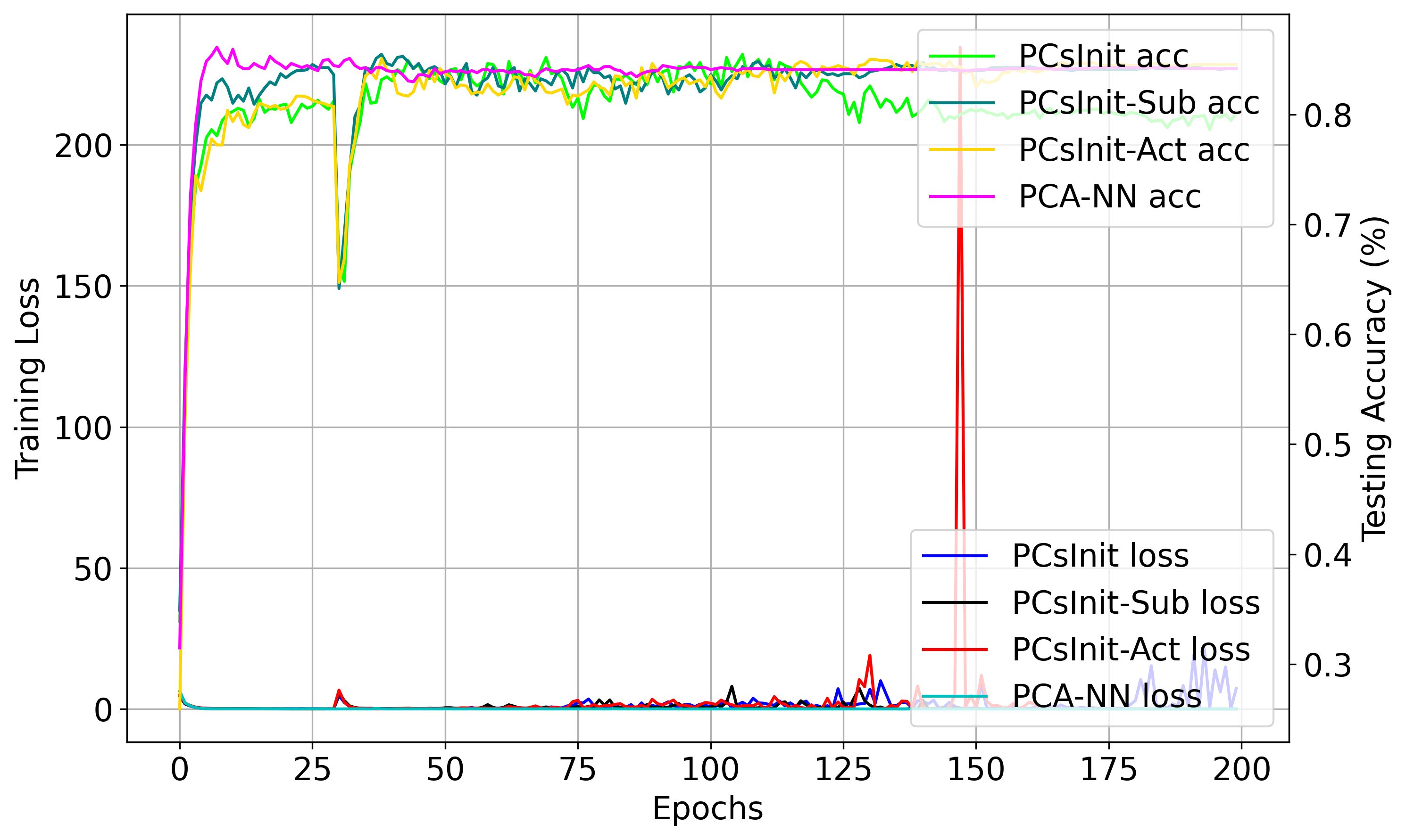}
    \vspace{-4mm}
    \caption{Micromass dataset with He initialization}
    \label{fig-micromass-he}
\end{figure}

\begin{figure}[!ht]
    \centering
    \includegraphics[width=0.7\linewidth]{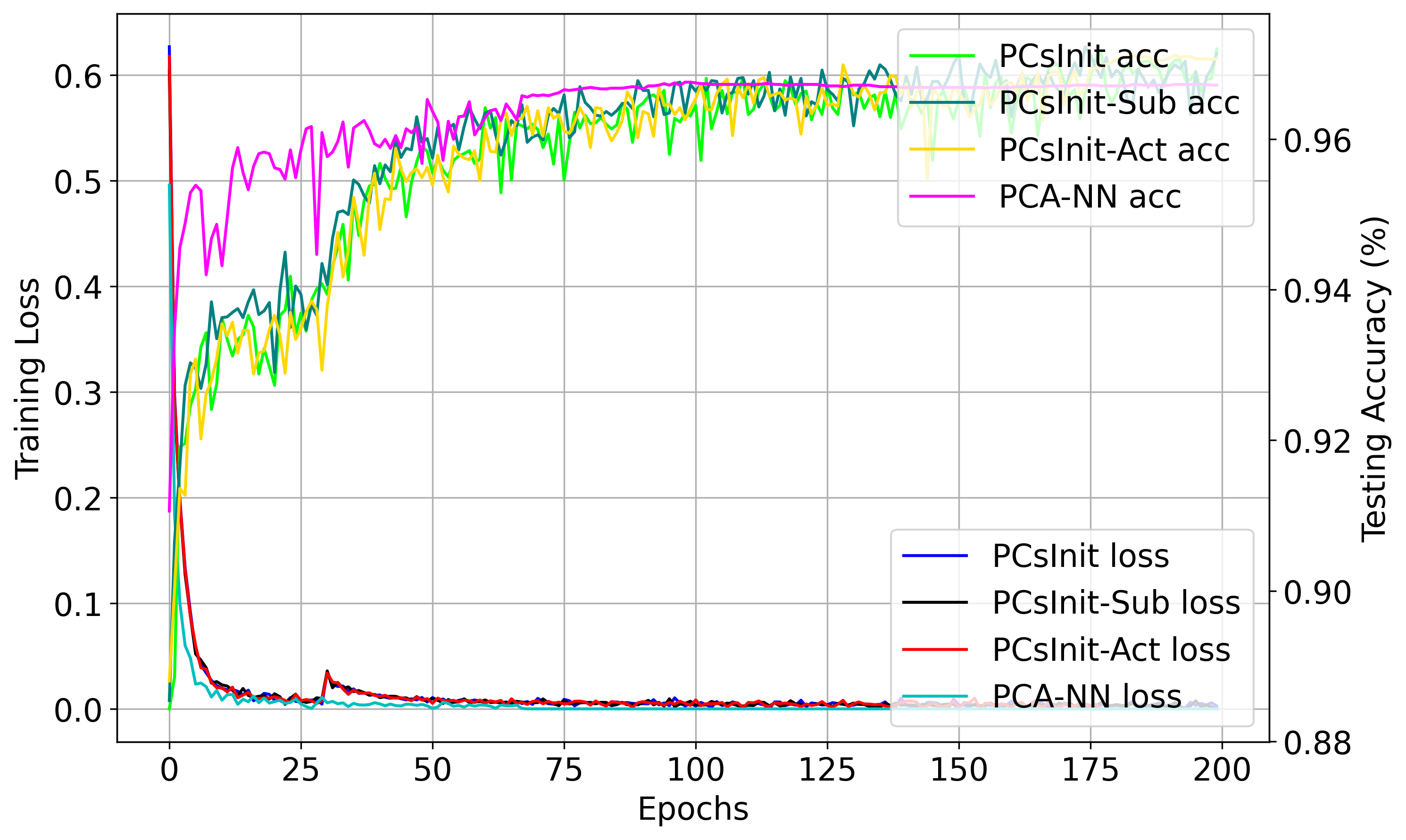}
    \vspace{-4mm}
    \caption{MNIST dataset with He initialization}
    \label{fig-mnist-noisy-he}
\end{figure}

\begin{figure}[!ht]
    \centering
    \includegraphics[width=0.7\linewidth]{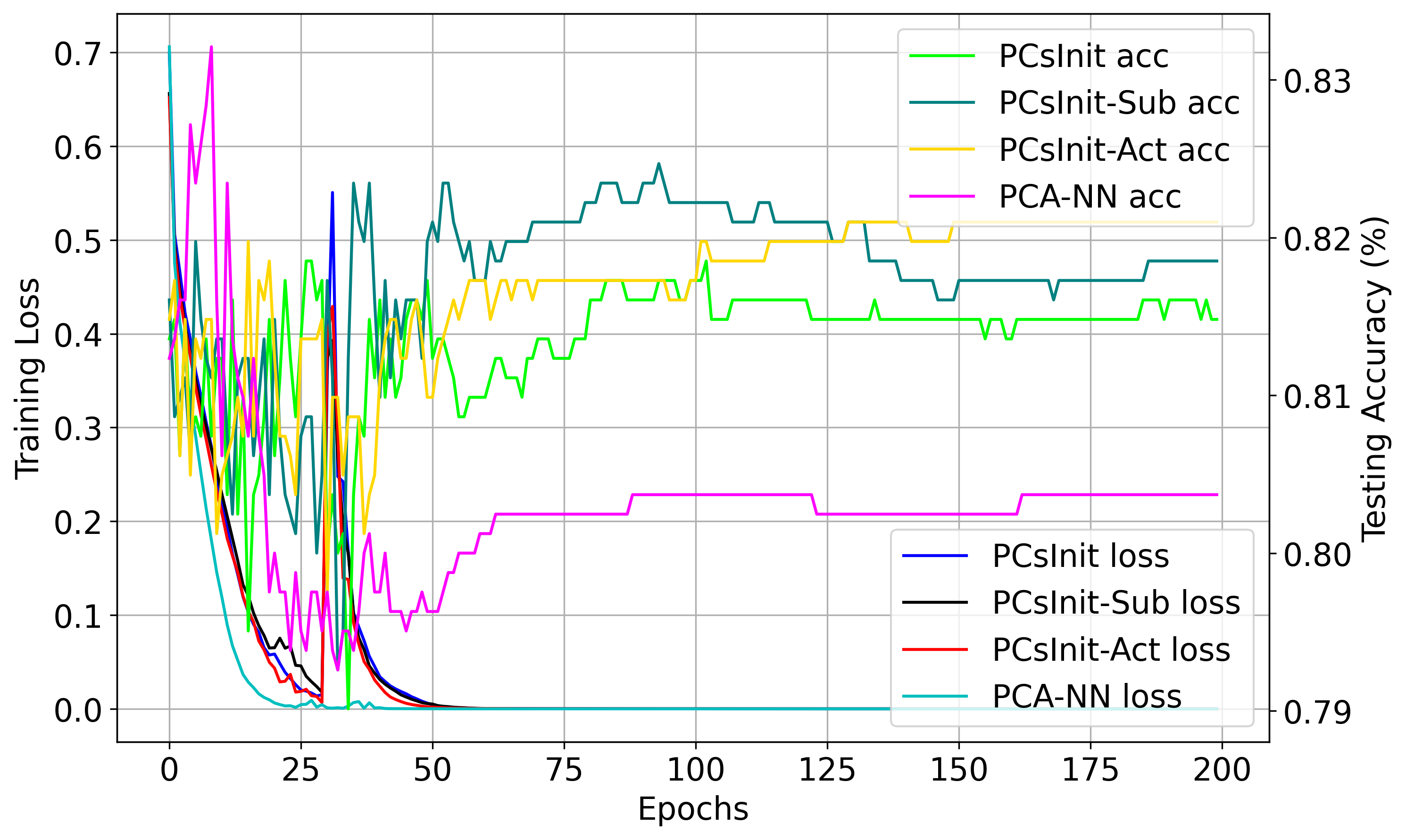}
    \caption{Heart dataset with He initialization}
    \label{fig-heart-he}
\end{figure}

\begin{figure}[!ht]
    \centering
    \includegraphics[width=0.7\linewidth]{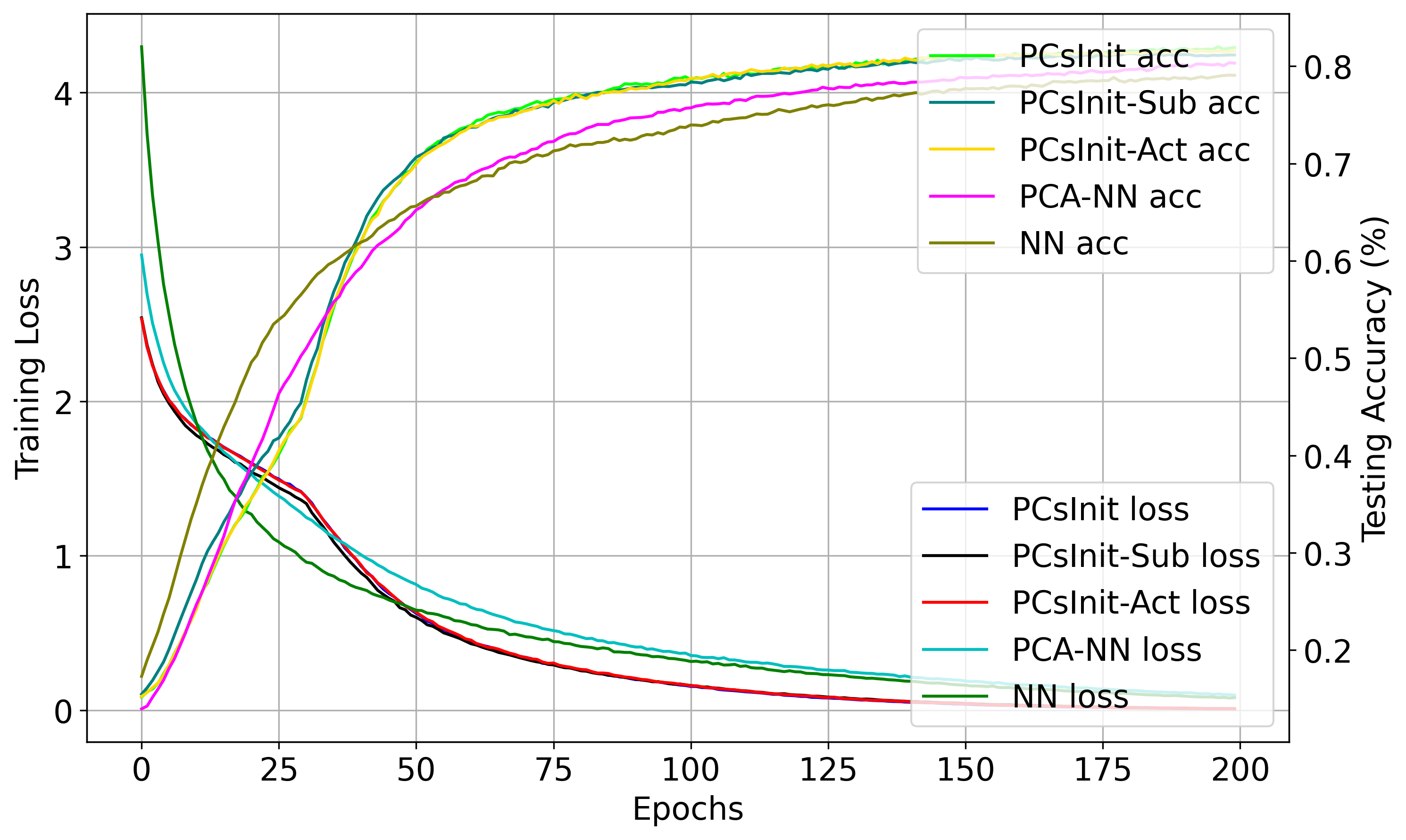}
    \caption{HTAD dataset with He initialization}
    \label{fig-htad-he}
\end{figure}

\begin{figure}[!ht]
    \centering
    \includegraphics[width=0.7\linewidth]{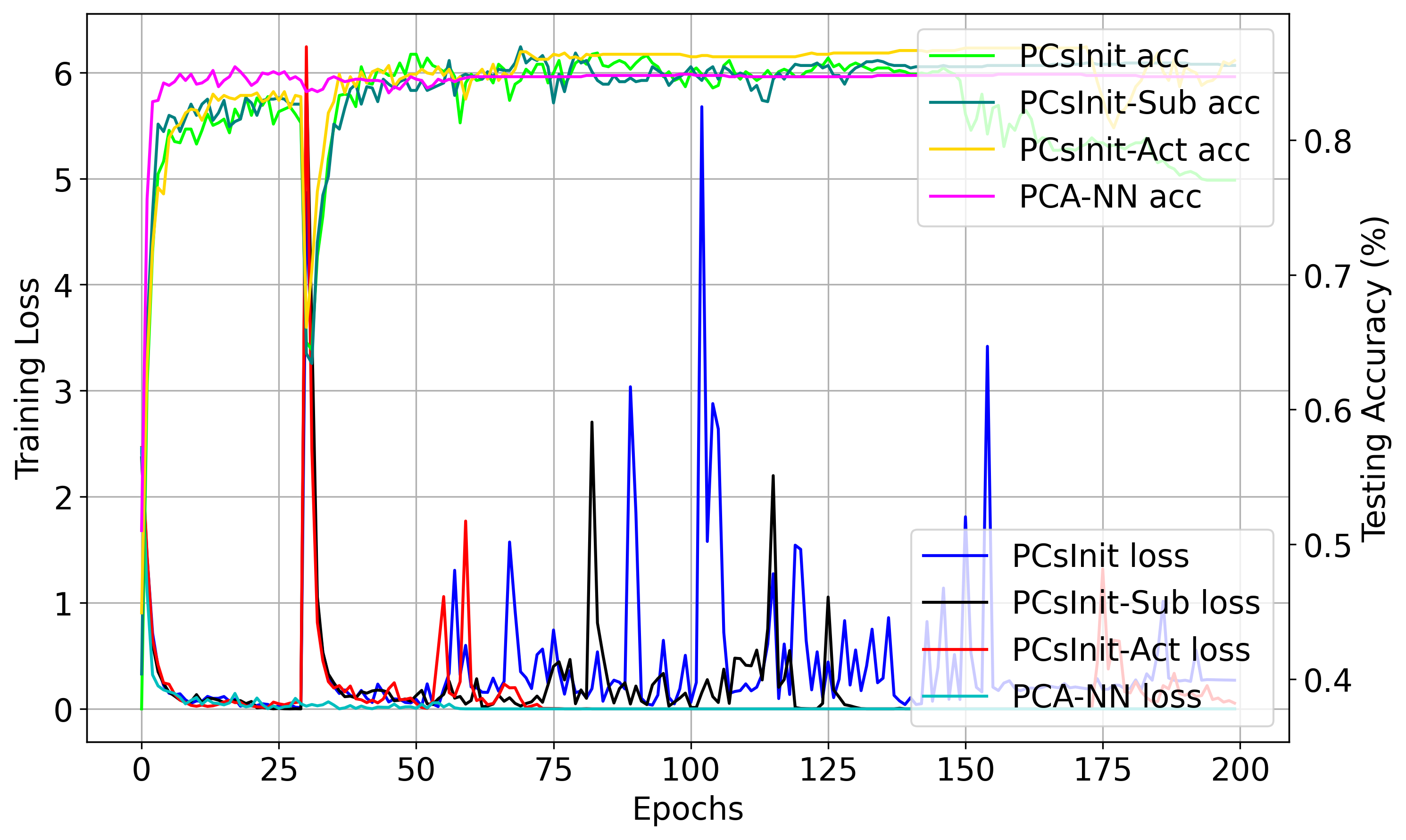}
    \caption{Micromass dataset with Xavier initialization}
    \label{fig-micromass-xavier}
\end{figure}

\begin{figure}[!ht]
    \centering
    \includegraphics[width=0.7\linewidth]{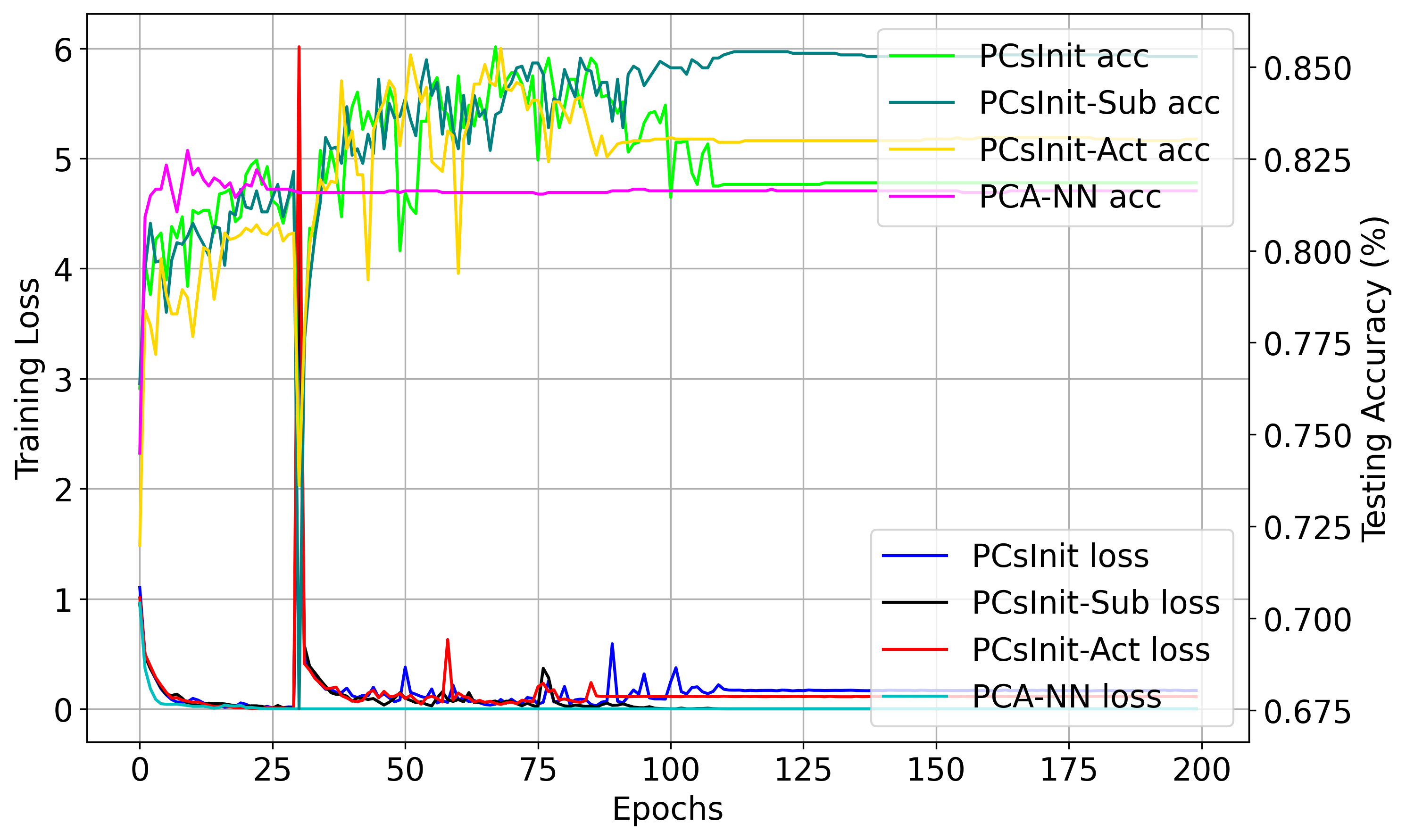}
    \caption{Parkinson dataset with Xavier initialization}
    \label{fig-parkinson-xavier}
\end{figure}

\begin{figure}[!ht]
    \centering
    \includegraphics[width=0.7\linewidth]{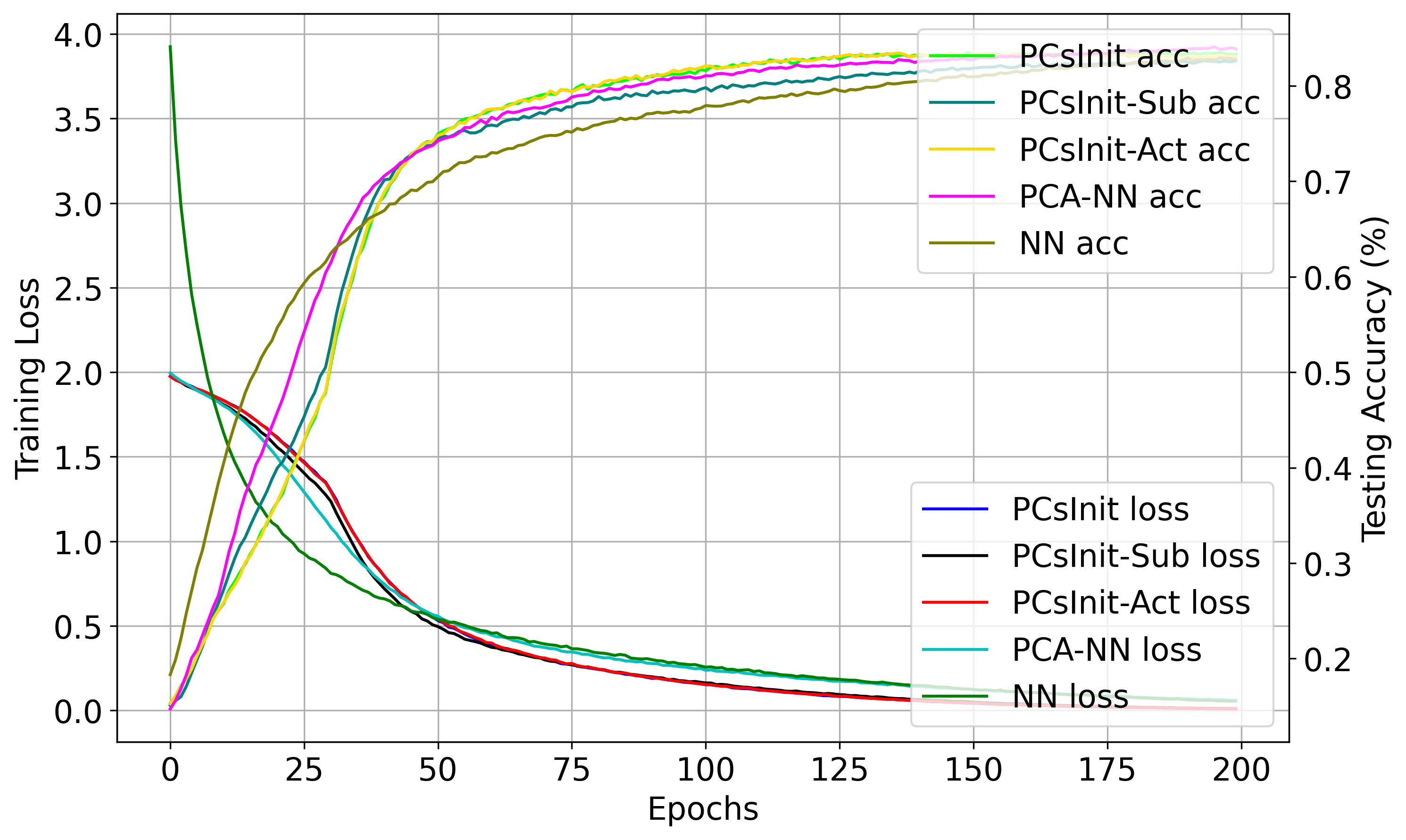}
    \caption{HTAD dataset with Xavier initialization}
    \label{fig-htad-xavier}
\end{figure}

\begin{figure}[!ht]
    \centering
    \includegraphics[width=0.7\linewidth]{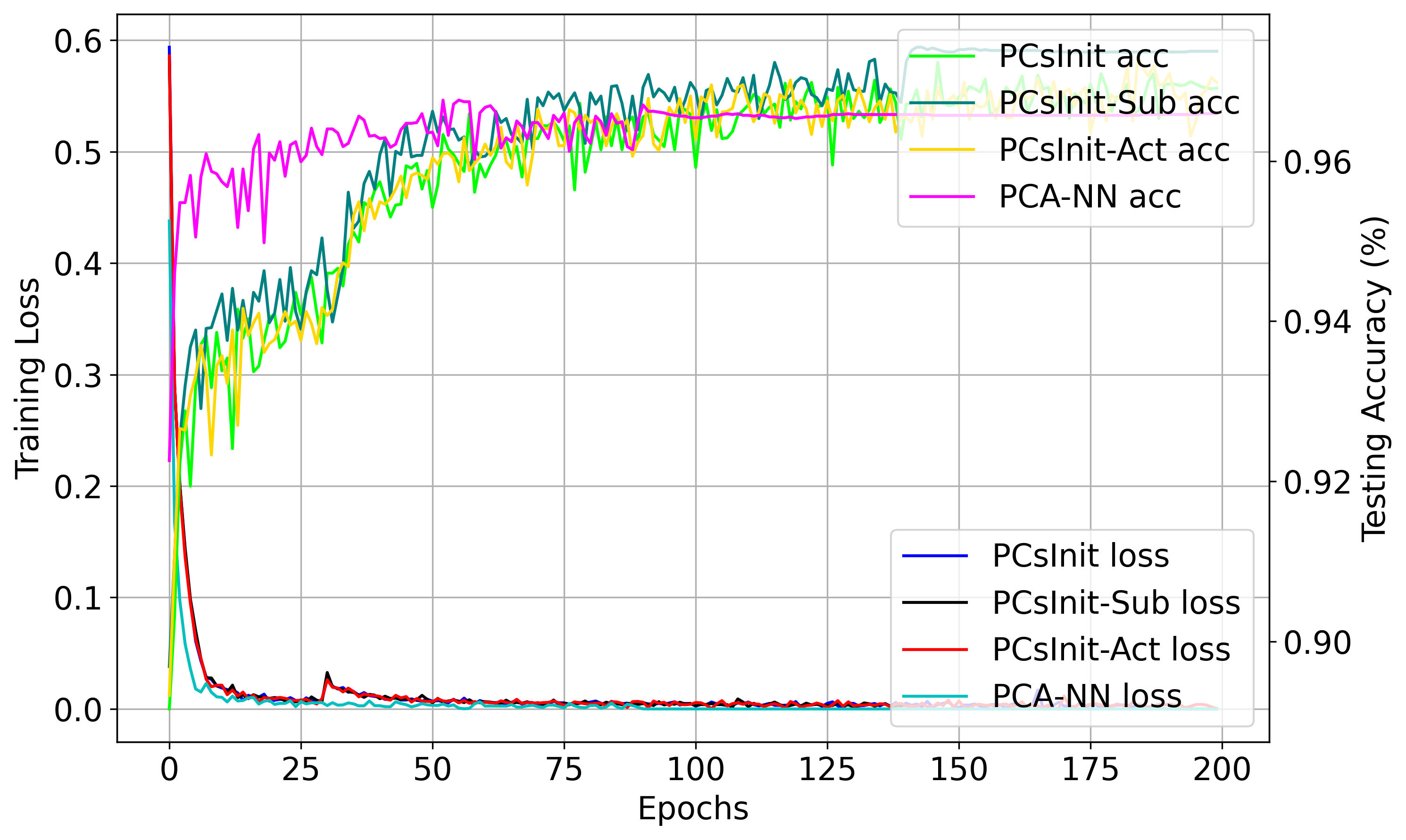}
    \caption{MNIST dataset with Xavier initialization}
    \label{fig-mnist-noisy-xavier}
\end{figure}

\begin{figure}[!ht]
    \centering
    \includegraphics[width=0.7\linewidth]{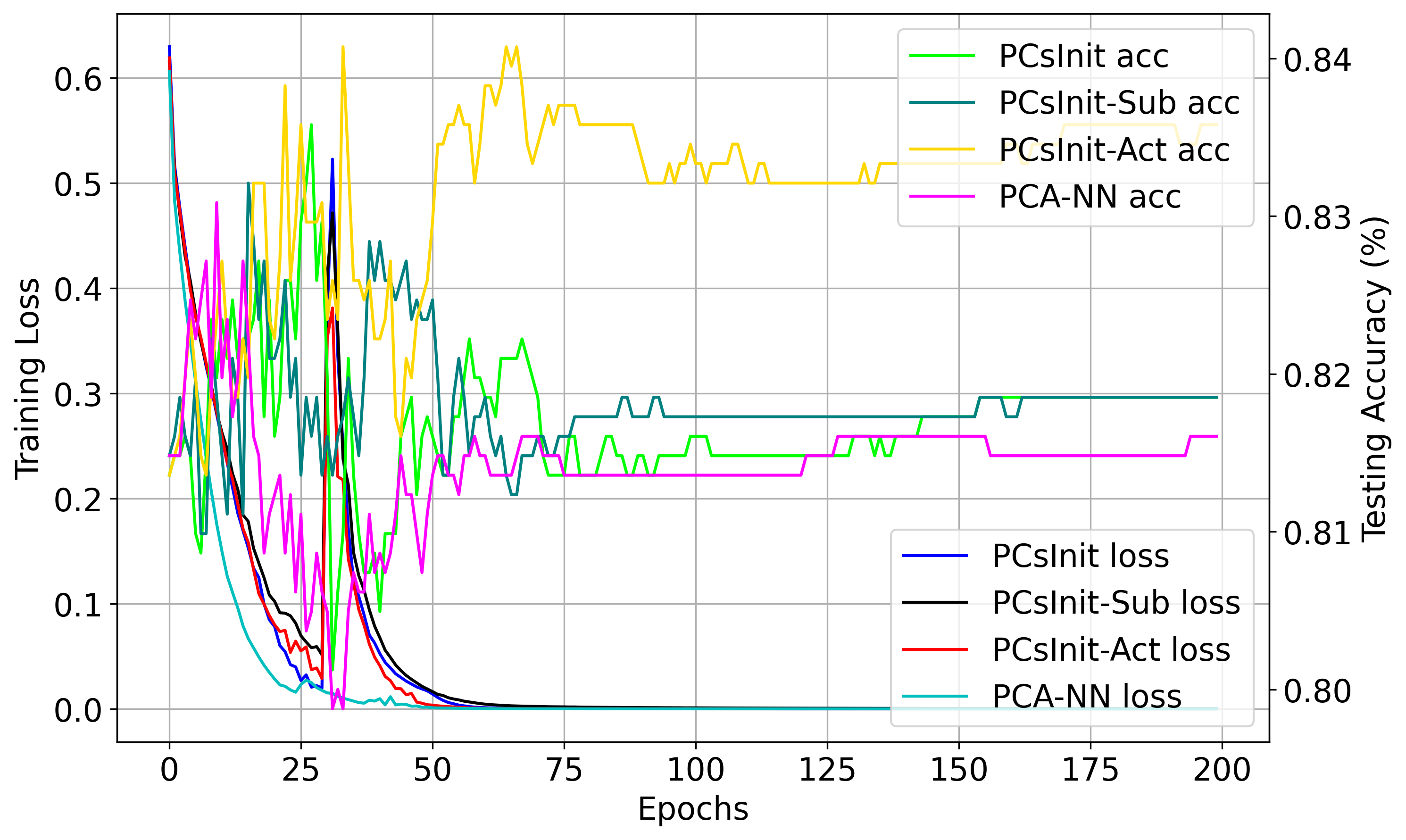}
    \caption{Heart dataset with Xavier initialization}
    \label{fig-heart-xavier}
\end{figure}

\begin{figure}[!ht]
    \centering
    \includegraphics[width=0.7\linewidth]{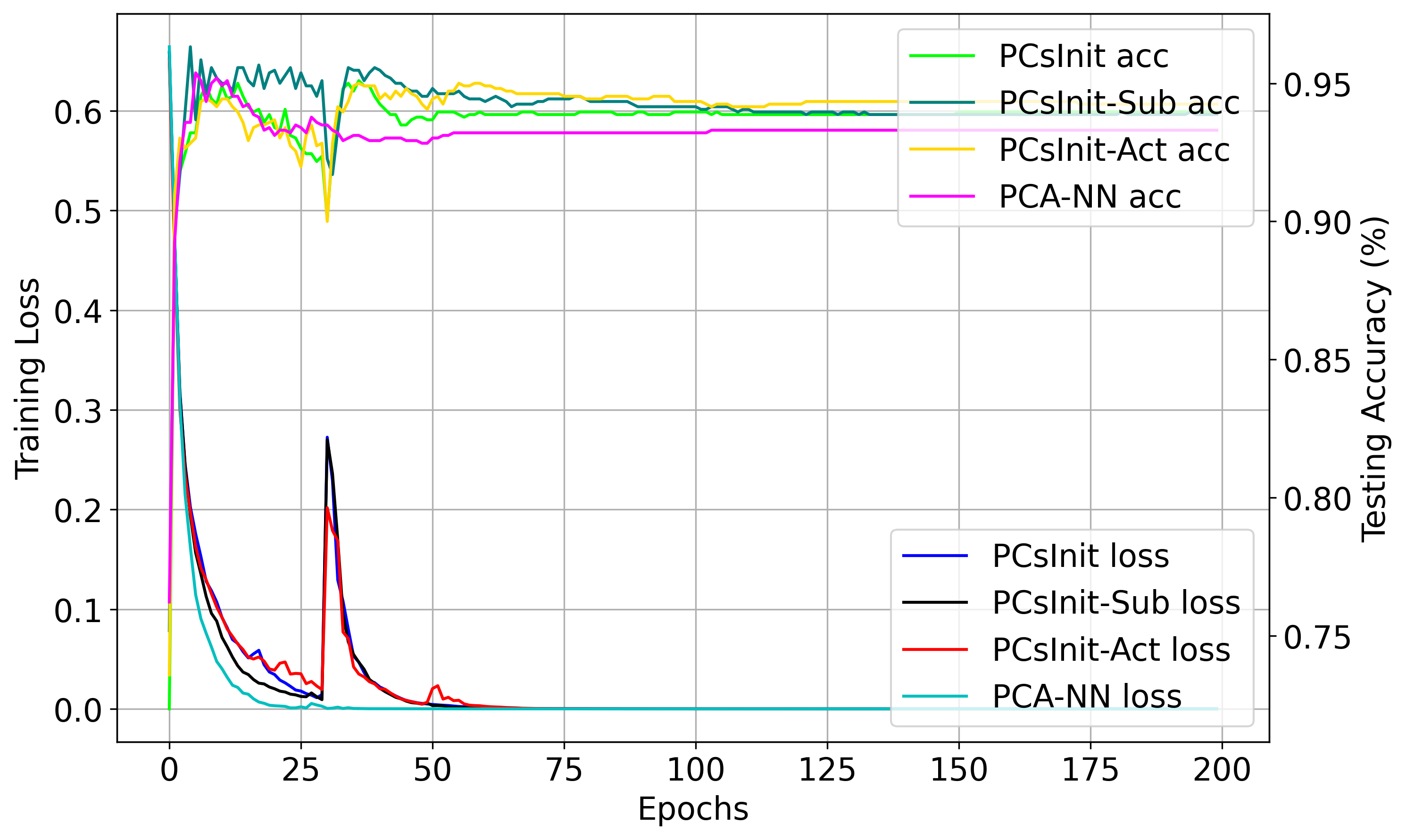}
    \caption{Ionosphere dataset with Xavier initialization}
    \label{fig-ionosphere-xavier}
\end{figure}


\begin{figure}[!ht]
    \centering
    \includegraphics[width=0.7\linewidth]{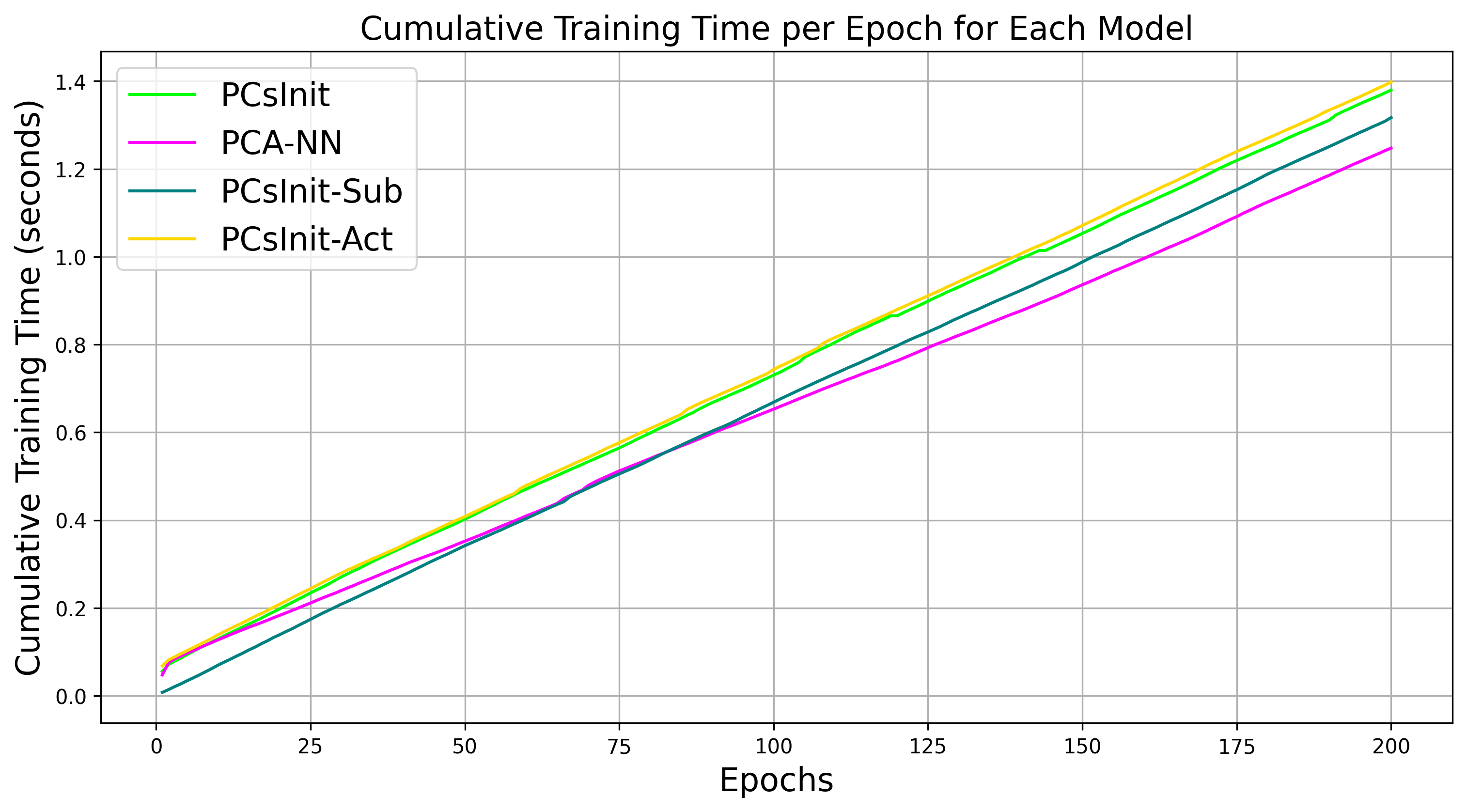}
    \caption{Running time on the Heart dataset with He initialization}
    \label{fig-time-heart-he}
\end{figure}

\subsection{Proof of Theorem 2.1}
\textbf{Statement.} Consider a single-layer linear neural network regression model with centered input data $X \in \mathbb{R}^{d \times n}$ (where each column represents a data point), target labels ${Y} \in \mathbb{R}^{1 \times n}$, and weight vector $W \in \mathbb{R}^{d \times 1}$. Let $J(W) = \frac{1}{2} \|W^T X - {Y}\|^2$ be the Mean Squared Error (MSE) loss function, whose Hessian matrix with respect to $W$ is $H$.
Suppose that $r$ principal components are used, $W_r$ is the matrix that consists of the selected eigenvectors of $X$, and let $Z = W_r^T X$. Then,
the Hessian of $J(W_r)$  with respect to $W_r$ has a condition number $\kappa(H_r)$ that satisfies:
\begin{equation}
    \kappa(H_r) \leq \kappa(H).
\end{equation}

\begin{proof}
It is established that the steepest descent method's convergence rate is significantly impacted by the condition number of the Hessian matrix and that convergence can be significantly slow when the condition number of the Hessian matrix is large \cite{ackleh2009classical}. For a single-layer linear regression problem with the output defined as $\hat{Y} = W^T X$, the Mean Squared Error (MSE) loss function is given by:

$$\qquad J(W) = \frac{1}{2} \|W^T X - Y\|^2.$$

The gradient of the loss function with respect to $W$ is:

$$\qquad \nabla J(W) = X (W^T X - Y)^T = X (X^T W - Y^T).$$

Therefore, the Hessian matrix is:

\begin{equation}
    \qquad H = \nabla^2 J(W) = X X^T. \label{equation-JW}
\end{equation}

The condition number of the Hessian matrix remains defined as the ratio of its largest eigenvalue ($\lambda_{max}$) to its smallest positive eigenvalue ($\lambda_{min}$):

$$\qquad \kappa(H) = \frac{\lambda_{max}}{\lambda_{min}}.$$




Since $W_r \in \mathbb{R}^{d \times r}$ be the matrix formed by the first $r$ eigenvectors of $X X^T$, the projected data is $Z = W_r^T X \in \mathbb{R}^{r \times n}$.
Therefore, the prediction is $W_r^T Z = W_r^T W_r^T X$. Based on \ref{equation-JW}, we see that the Hessian of the loss function with respect to $W_r$ in the reduced space would be $H_r = Z Z^T = (W_r^T X) (W_r^T X)^T = W_r^T X X^T W_r$.

Let $V \in \mathbb{R}^{d \times d}$ be the semi-orthonormal matrix whose columns are the eigenvectors of $X X^T$, ordered by decreasing eigenvalue. 
Since $V$ diagonalizes $X X^T$ such that $V^T X X^T V = \Lambda = \text{diag}(\lambda_1, \lambda_2, ..., \lambda_d)$ with $\lambda_1 \geq \lambda_2 \geq ... \geq \lambda_d \geq 0$, then $W_r^T X X^T W_r = \Lambda_r = \text{diag}(\lambda_1, \lambda_2, ..., \lambda_r)$.
By discarding the smallest eigenvalues (corresponding to directions of low variance), PCA effectively works with a Hessian $H_r$ whose condition number is:

$$\qquad \kappa(H_r) = \frac{\lambda_1}{\lambda_r}$$

Since $\lambda_r \geq \lambda_{min}$ (where $\lambda_{min}$ is the smallest positive eigenvalue of the original $H = X X^T$), we have $\kappa(H_r) \leq \kappa(H)$. Furthermore, if some of the smallest eigenvalues of $H$ are close to zero, their removal through dimensionality reduction via PCA can significantly reduce the condition number of the effective Hessian in the reduced space, leading to more stable and potentially faster convergence of optimization algorithms. The PCsInit method leverages this by initializing the weights in a way that aligns with the principal components, effectively operating in a space where the Hessian is better conditioned.
\end{proof}

\subsection{Proof for theorem 2.2}
\textbf{Statement.} Let $x \in \mathbb{R}^{d}$ be the input vector, and let $W_r \in \mathbb{R}^{d \times r}$ be the weight matrix of the first layer, where $r \leq d$ is the number of principal components used. The columns of $W_r$ are the principal component vectors. Assuming that the bias term is zero, i.e., the output of the first layer is $\qquad h^1 = W_r^T x$.  
Then, the Lipschitz constant for the first layer in PCsInit is:
    \begin{equation*}
 L_1 = \sigma_{max}(W_r) = ||W_r||.
    \end{equation*}
    where $||.||$ is the spectral norm.
\begin{proof}
Let's consider the difference in the output of the first layer for two different inputs, $x$ and $y$:
    
    $$\qquad h^1(x) - h^1(y) = W_r^T x - W_r^T y = W_r^T (x - y).$$
    
Now, take the norm of both sides:
\begin{align*}    
    \qquad ||h^1(x) - h^1(y)|| &= ||W_r^T (x - y)||\\
    &\leq ||W_r^T|| \cdot ||x - y||.
\end{align*}
It is known that the spectral norm of a matrix is equal to its largest singular value \cite{nguyen2022matrix}. Since the singular values of $W_r^T$ are the same as the singular values of $W_r$, we can also say that $||W_r^T|| = ||W_r||$.
 Let $\sigma_{max}(W_r)$ be the largest singular value of $W_r$. Then:
    
    $$\qquad ||h^1(x) - h^1(y)|| \leq \sigma_{max}(W_r) \cdot ||x - y||.$$
    
    Therefore, the Lipschitz constant for the first layer in PCsInit is:
    
    $$\qquad L_1 = \sigma_{max}(W_r) = ||W_r||.$$
    
\end{proof} 

\subsection{Proof for theorem 2.3}

\textbf{Statement.} Consider the first layer of a neural network with the weight matrix $W_r \in \mathbb{R}^{d \times r}$
consists of the first $r$ principal components (PCs) of the input data, forming semi-orthonormal rows, such
that $W_r = Q$, where $Q \in \mathbb{R}^{d \times r}$ has semi-orthonormal rows ($Q Q^T = I_d$). Following this linear layer,
an element-wise Lipschitz continuous activation function $\sigma : \mathbb{R} \rightarrow \mathbb{R}$ with Lipschitz constant $L_\sigma$ is
applied. Then, the Lipschitz constant $L_1$ of the first layer operation $f_1(x) = \sigma(W_r^T x)$ with respect
to the $L_2$ norm is $L_1 = L_\sigma$.

\begin{proof}
Let $f_1: \mathbb{R}^{d} \rightarrow \mathbb{R}^{r}$ be the operation of the first layer, defined as $f_1(x) = \sigma(W_r^T x)$, where $\sigma$ is applied element-wise. We aim to find the Lipschitz constant $L_1$ such that for all $x, y \in \mathbb{R}^{d}$:

$$||f_1(x) - f_1(y)||_2 \le L_1 ||x - y||_2.$$

Substituting the definition of $f_1(x)$:

$$||\sigma(W_r^T x) - \sigma(W_r^T y)||_2 \le L_1 ||x - y||_2.$$

Let $a = W_r^T x$ and $b = W_r^T y$. Then the inequality becomes:

$$||\sigma(a) - \sigma(b)||_2 \le L_1 ||x - y||_2.$$

Since $\sigma$ is an element-wise Lipschitz continuous activation function with Lipschitz constant $L_\sigma$, for each component $i \in \{1, ..., r\}$:

$$|\sigma(a_i) - \sigma(b_i)| \le L_\sigma |a_i - b_i|.$$

Squaring both sides and summing over all components:

$$\sum_{i=1}^{r} (\sigma(a_i) - \sigma(b_i))^2 \le L_\sigma^2 \sum_{i=1}^{r} (a_i - b_i)^2.$$

Taking the square root of both sides yields:

$$||\sigma(a) - \sigma(b)||_2 \le L_\sigma ||a - b||_2.$$

Substituting back $a = W_r^T x$ and $b = W_r^T y$:

$$||\sigma(W_r^T x) - \sigma(W_r^T y)||_2 \le L_\sigma ||W_r^T x - W_r^T y||_2.$$

By the linearity of matrix multiplication:

$$||\sigma(W_r^T x) - \sigma(W_r^T y)||_2 \le L_\sigma ||W_r^T (x - y)||_2.$$

Given that the weight matrix $W_r$ is formed by the first $r$ principal components of the input data, forming semi-orthonormal rows, we have $W_rW_r^T=I$.
Therefore, the largest singular value (which is the spectral norm) is 1.
Substituting this back into the Lipschitz inequality:
\begin{align*}
||\sigma(W_r^T x) - \sigma(W_r^T y)||_2 &\le L_\sigma (1) ||x - y||_2\\
\Rightarrow ||\sigma(W_r^T x) - \sigma(W_r^T y)||_2 &\le L_\sigma ||x - y||_2.    
\end{align*}

Thus, the Lipschitz constant of the first layer $f_1(x) = \sigma(W_r^T x)$ is $L_1 = L_\sigma$.
    
\end{proof}

\subsection{Proof of theorem 2.4}
\textbf{Statement.} Assume that $\tilde{x} = x + \eta,$ where $\tilde{x}$ is the noisy input, $x$ is the clean input, and $\eta$ is
the noise vector. In addition, assume also that $\eta \sim \mathcal{N}(0, \sigma^2I)$, i.e., the noise follows a Gaussian
distribution with zero mean and covariance matrix $\sigma^2I$. Here, $\sigma^2 \in \mathbb{R}^+$ and $I$ is the identity matrix.
Next, let the eigenvalues of $X^T X$ be $\lambda_1, \lambda_2, ..., \lambda_r$. Then, for PCsInit, the noise propagated after the
first layer is $W_r^T \eta$ follows Gaussian distribution with mean 0 and its covariance matrix is a diagonal
matrix with entries $\sigma^2 \lambda_1, \sigma^2 \lambda_2, ..., \sigma^2 \lambda_r$.

\begin{proof}
The output of the first layer is
\begin{align}
h^1 & = W_r^T \tilde{x}\\
    & = W_r^T (x + \eta) = W_r^T x + W_r^T \eta.
\end{align}

Since $\eta \sim \mathcal{N}(0, \sigma^2I)$, the transformed noise $W_r^T \eta$ will also be Gaussian with mean:
\begin{equation}
    \mathbb{E}[W_r^T \eta] = W_r^T \mathbb{E}[\eta] = W_r^T 0 = 0.
\end{equation}
and covariance matrix:
\begin{align}
    Cov(W_r^T \eta) &= \mathbb{E}[(W_r^T \eta-0)(W_r^T \eta-0)^T] = W_r^T \mathbb{E}[\eta \eta^T] W_r \\
    &= W_r^T (\sigma^2 I) W_r = \sigma^2 W_r^T W_r.
\end{align}

Therefore, $W_r^T \eta \sim \mathcal{N}(0, \sigma^2 W_r^T W_r)$.

Since $W_r$ comes from the selected $r$ eigenvalues - eigenvectors pair of PCA, $W_r^T W_r$ is a diagonal matrix
 with the eigenvalues of $X^T X$ on the diagonal.  Therefore, the covariance matrix of $W_r^T \eta$ is a diagonal matrix with entries $\sigma^2 \lambda_1, \sigma^2 \lambda_2, ..., \sigma^2 \lambda_r$.
\end{proof}

\subsection{Proof of theorem 2.5}
\textbf{Statement.}     Consider a neural network where the first layer's weight matrix, \(W_r \in \mathbb{R}^{d\times r}\), is initialized as in PCsInit, and therefore \(W_r = Q\), where \(Q \in \mathbb{R}^{d \times r}\) is semi-orthonormal. 
     Suppose that the input to the network is corrupted by additive white noise, i.e.,  \(\tilde{x} = x + \eta\), where \(x \in \mathbb{R}^{d}\) is the clean input signal and \(\eta \sim \mathcal{N}(0, \sigma^2 I)\) is the additive white noise.
     Then, the norm of noise component after the first layer is preserved, i.e.,
        \(   ||W_r^T\eta|| = ||\eta||.   \)

     Also, let \(f: \mathbb{R}^{d} \rightarrow \mathbb{R}^{r}\) be the neural network function, decomposed into layers \(f = f_L \circ f_{L-1} \circ ... \circ f_1\), where \(f_i\) represents the \(i\)-th layer,
     \(f(x)\) is the clean output, and \(\tilde{f}(\tilde{x})\) is the noisy output.
    Further assume that each subsequent layer \(f_i\) for \(i = 2, ..., L\) is \(L_i\)-Lipschitz continuous. Then,
        \[   ||\tilde{f}(\tilde{x}) - f(x)|| \le \left( \prod_{i=2}^{L} L_i \right) ||\eta||.   \]

\begin{proof}
     For PCsInit, the output of the first layer with noisy input is:
        \[   \tilde{h}_1 = f_1(\tilde{x}) = W_r^T (x + \eta) = W_r^T x + W_r^T \eta = h_1 + W_r^T \eta   \]
     Since \(W_r = Q\) (where $Q \in \mathbb{R}^{d \times r}$ is semi-orthonormal, meaning $Q^T Q = I_r$, so $W_r^T W_r = I_r$), and thus $W_r^T$ (an $r \times d$ matrix) has semi-orthonormal rows ($W_r^T (W_r^T)^T = W_r^T W_r = I_r$), so \(||W_r^T|| = 1\). This implies that the noise component's norm is preserved:
        \[   ||W_r^T \eta|| = ||\eta||   \]

     Using the Lipschitz property of subsequent layers, we can bound the noise propagation:
\begin{align*}
    ||\tilde{f}(\tilde{x}) - f(x)|| 
        &= ||f_L(f_{L-1}(...f_1(\tilde{x})))-f_L(f_{L-1}(...f_1(x)))||\\
        &\le L_LL_{L-1}...L_2||f_1(\tilde{x})-f_1(x)||\\
        &\le \left( \prod_{i=2}^{L} L_i \right) ||\eta||         
\end{align*}
        
\end{proof}
\subsection{Proof of theorem 2.6}
\textbf{Statement.} For any layer $\ell > 1$, let $\rho^{\ell}$ is the activation function, $W^{\ell}$ is the weight matrix, and $b^{\ell}$ is the bias vector. Then, the output at layer $\ell$ is: $h^{\ell} = \rho^{\ell}(W^{\ell} h^{\ell-1} + b^{\ell}).$ Next, let $\eta^{\ell}$ represents the noise in the output of layer $\ell$ and assume also that $\rho^{\ell}$ is Lipschitz continuous with Lipschitz constant $L_{\ell}$, i.e.
$\qquad ||h^{\ell} - \tilde{h}^{\ell}|| \leq L_{\ell} ||W^{\ell}|| \cdot ||\eta^{\ell-1}||,$
where $\tilde{h}^{\ell}$ is the clean output and $h^{\ell}$ is the noisy output.
Then, the bound for the noise propagated to the second layer specifically, which this isolates the influence of the first layer initialization of PCsInit compared to other techniques is
\begin{equation}
    \qquad ||\eta^2|| \leq L_2 ||W^2|| \cdot ||W^1|| \cdot ||\eta^0||.
\end{equation}

In addition, the general noise bound for any layer $\ell > 1$ is:
\begin{equation}
||\eta^{\ell}|| \leq \left[\prod_{i=2}^l(L_i||W^i||)\right]\cdot ||W^1|| \cdot ||\eta^0||,
\end{equation}    

\begin{proof}
    
Substituting the noisy input from the previous layer gives 
\begin{equation}
    h^{\ell} = \rho^{\ell}(W^{\ell} (h^{\ell-1} + \eta^{\ell-1}) + b^{\ell}) = \rho^{\ell}(W^{\ell} h^{\ell-1} + b^{\ell} + W^{\ell} \eta^{\ell-1})
\end{equation}


We first examine the \textbf{noise propagation to layer 2}:

Let $\eta^0 = \eta$ be the noise added to the input: $\tilde{x} = x + \eta$.
The output of the first layer with noisy input is:
    
    $$\qquad h^1 = W^1 \tilde{x} = W^1 (x + \eta) = W^1 x + W^1 \eta$$
    
The output of the first layer with clean input is:
    
    $$\qquad \tilde{h}^1 = W^1 x$$.
    
Therefore, the noise at the output of layer 1 is:
\begin{equation}\label{eq-noise-layer1}
    \qquad \eta^1 = h^1 - \tilde{h}^1 = W^1 \eta
\end{equation}
    
The output of the second layer with noisy input from the first layer is:
    
    $$\qquad h^2 = \rho^2(W^2 h^1 + b^2) = \rho^2(W^2 (W^1 x + W^1 \eta) + b^2).$$
    
The output of the second layer with clean input from the first layer is:
    
    $$\qquad \tilde{h}^2 = \rho^2(W^2 \tilde{h}^1 + b^2) = \rho^2(W^2 (W^1 x) + b^2).$$
    
Therefore, the noise at the output of layer 2 is:
    
    $$\qquad \eta^2 = h^2 - \tilde{h}^2 = \rho^2(W^2 h^1 + b^2) - \rho^2(W^2 \tilde{h}^1 + b^2) = \rho^2(W^2 (W^1 x + W^1 \eta) + b^2) - \rho^2(W^2 (W^1 x) + b^2)$$

 Using the Lipschitz property of the activation function $\rho^2$ (with Lipschitz constant $L_2$), we have:
    
    $$\qquad ||\eta^2|| = ||h^2 - \tilde{h}^2|| \leq L_2 ||W^2 (h^1 - \tilde{h}^1)|| = L_2 ||W^2 \eta^1|| \leq L_2 ||W^2|| \cdot ||\eta^1||$$
    
 Substituting the noise at the output of layer 1 in \ref{eq-noise-layer1}:
    
    $$\qquad ||\eta^2|| \leq L_2 ||W^2|| \cdot ||W^1 \eta|| \leq L_2 ||W^2|| \cdot ||W^1|| \cdot ||\eta||$$
    
  The difference (noise) in the output of layer $\ell$ is $\eta^{\ell} = h^{\ell} - \tilde{h}^{\ell}$ and we can apply recursive application to derive its bound. Recall that for layer 1, we already established that:

$$\qquad ||\eta^1|| = ||h^1 - \tilde{h}^1|| \leq ||W^1|| \cdot ||\eta^0||$$

Next, for layer 2:

$$\qquad ||\eta^2|| \leq L_2 ||W^2|| \cdot ||W^1|| \cdot ||\eta^0||$$
    
Now, for layer 3:
    
    $$\qquad ||\eta^3|| = ||h^3 - \tilde{h}^3|| \leq L_3 ||W^3|| \cdot ||\eta^2||.$$
    
Hence, substiting the bound for $||\eta^2||$ we have:
    
    $$\qquad ||\eta^3|| \leq L_3 ||W^3|| \cdot (L_2 ||W^2|| \cdot ||W^1|| \cdot ||\eta^0||) = L_3 ||W^3|| \cdot L_2 ||W^2|| \cdot ||W^1|| \cdot ||\eta^0||$$
    
By continuing this pattern, we get for any layer $\ell > 1$:
    
    $$\qquad ||\eta^{\ell}|| \leq (L_{\ell} ||W^{\ell}||) \cdot (L_{\ell-1} ||W^{\ell-1}||) \cdot ... \cdot (L_2 ||W^2||) \cdot ||W^1|| \cdot ||\eta^0||$$
    
 For the output layer (layer $L$), the bound becomes:
    
    $$\qquad ||\eta^L|| \leq (L_L ||W^L||) \cdot (L_{L-1} ||W^{L-1}||) \cdot ... \cdot (L_2 ||W^2||) \cdot ||W^1|| \cdot ||\eta^0||$$

\end{proof}

\end{document}